\def\year{2020}\relax
\documentclass[letterpaper]{article} 
\usepackage{aaai20}  
\usepackage{times}  
\usepackage{helvet} 
\usepackage{courier}  
\usepackage[hyphens]{url}  
\usepackage{graphicx} 
\usepackage{graphicx}  
\usepackage{amsmath,amsfonts,bm}

\def\eqref#1{equation~\ref{#1}}

\def\1{\bm{1}}

\def\rz{{\textnormal{z}}}

\def\vzero{{\bm{0}}}

\def\vmu{{\bm{\mu}}}
\def\vtheta{{\bm{\theta}}}
\def\vtau{{\bm{\tau}}}
\def\vzero{{\bm{0}}}
\def\vI{{\bm{I}}}
\def\va{{\bm{a}}}

\def\vphi{{\bm{\phi}}}

\def\vo{{\bm{o}}}

\def\vs{{\bm{s}}}

\def\vu{{\bm{u}}}
\def\vv{{\bm{v}}}
\def\vw{{\bm{w}}}
\def\vx{{\bm{x}}}

\def\vz{{\bm{z}}}

\def\mSigma{{\bm{\Sigma}}}

\DeclareMathAlphabet{\mathsfit}{\encodingdefault}{\sfdefault}{m}{sl}
\SetMathAlphabet{\mathsfit}{bold}{\encodingdefault}{\sfdefault}{bx}{n}

\def\sA{{\mathbb{A}}}

\def\sG{{\mathbb{G}}}

\def\sI{{\mathbb{I}}}

\def\sO{{\mathbb{O}}}

\def\sS{{\mathbb{S}}}
\def\sT{{\mathbb{T}}}

\def\sX{{\mathbb{X}}}

\def\sZ{{\mathbb{Z}}}

\newcommand{\E}{\mathbb{E}}

\newcommand{\KL}{D_{\mathrm{KL}}}

\usepackage{multicol}
\usepackage{subcaption,graphicx}
\usepackage{dblfloatfix} 
\usepackage{hyperref}
\usepackage{xcolor}
\usepackage{url}
\usepackage{graphicx}
\graphicspath{{images/}}
\usepackage{subcaption}
\usepackage{float}
\usepackage{algorithm}
\usepackage[noend]{algpseudocode}
\usepackage{tikz}
\usepackage{pgfplots}
\usepackage{color}

\usepackage{natbib}
\title{Variational Autoencoders for Opponent Modeling in Multi-Agent Systems}

\author{Georgios Papoudakis   \\
School of Informatics\\
The University of  Edinburgh\\
Edinburgh Centre for Robotics\\
\texttt{g.papoudakis@ed.ac.uk} \\
\And Stefano V. Albrecht \\
School of Informatics\\
The University of  Edinburgh\\
Edinburgh Centre for Robotics\\
\texttt{s.albrecht@ed.ac.uk} \\
\AND
}

\begin{document}
%
\maketitle

\newcommand{\fix}{\marginpar{FIX}}
\newcommand{\new}{\marginpar{NEW}}


\definecolor{blue_sns}{HTML}{1F77B4} 
\definecolor{orange_sns}{HTML}{FF7F00} 
\definecolor{green_sns}{HTML}{2CA02C} 
\definecolor{red_sns}{HTML}{D62628} 
\definecolor{background_sns}{HTML}{EAEAF2} 
\definecolor{sma2csa}{HTML}{E377C2} 
\definecolor{sma2cs}{HTML}{7F7F7F} 
\definecolor{sma2noback}{HTML}{17BECF} 

\newcommand{\smac}{\raisebox{2pt}{\tikz{\draw[blue_sns,solid,line width=1.2pt](0,0) -- (5mm,0);}}}
\newcommand{\omddpg}{\raisebox{2pt}{\tikz{\draw[green_sns,solid,line width=1.2pt](0,0) -- (5mm,0);}}}
\newcommand{\grover}{\raisebox{2pt}{\tikz{\draw[red_sns,solid,line width=1.2pt](0,0) -- (5mm,0);}}}
\newcommand{\omddpgnodisc}{\raisebox{2pt}{\tikz{\draw[orange_sns,solid,line width=1.2pt](0,0) -- (5mm,0);}}}
\newcommand{\smacsa}{\raisebox{2pt}{\tikz{\draw[sma2csa,solid,line width=1.2pt](0,0) -- (5mm,0);}}}
\newcommand{\smacs}{\raisebox{2pt}{\tikz{\draw[sma2cs,solid,line width=1.2pt](0,0) -- (5mm,0);}}}
\newcommand{\smanoback}{\raisebox{2pt}{\tikz{\draw[sma2noback,solid,line width=1.2pt](0,0) -- (5mm,0);}}}
\maketitle

\begin{abstract}
Multi-agent systems exhibit complex behaviors that emanate from the interactions of multiple agents in a shared environment. In this work, we are interested in controlling one agent in a multi-agent system and successfully learn to interact with the other agents that have fixed policies. Modeling the behavior of other agents (opponents) is essential in understanding the interactions of the agents in the system. By taking advantage of recent advances in unsupervised learning, we propose modeling opponents using variational autoencoders. Additionally, many existing methods in the literature assume that the opponent models have access to opponent's observations and actions during both training and execution. To eliminate this assumption, we propose a modification that attempts to identify the underlying opponent model using only local information of our agent, such as its observations, actions, and rewards. The experiments indicate that our opponent modeling methods achieve equal or greater episodic returns in reinforcement learning tasks against another modeling method.
\end{abstract}

\section{Introduction}
\label{sec:intro}

In recent years, several promising works \citep{mnih2015human, schulman2015trust, mnih2016asynchronous} have arisen in deep reinforcement learning (RL), leading to fruitful results in single-agent scenarios. In this work, we are interested in using single-agent RL in multi-agent systems, where we control one agent and the other agents (opponents) in the environment have fixed policies. The agent should be able to successfully interact with a diverse set of opponents as well as generalize to new unseen opponents. One effective way to address this problem is opponent modeling. The opponent models output specific characteristics of the opponents based on their trajectories. By successfully modeling the opponents, the agent can reason about opponents' behaviors and goals and adjust its policy to achieve optimal outcomes. There is a rich literature of modeling opponents in multi-agent systems \citep{albrecht2018autonomous}.



Several recent works have proposed learning opponent models using deep learning architectures \citep{he2016opponent,raileanu2018modeling,grover2018learning,rabinowitz2018machine}. In this work, we focus on learning opponent models using Variational Autoencoders (VAEs) \citep{kingma2013auto}. VAE are generative models that are commonly used for learning representations of data, and various works use them in RL for learning representations of the environment \citep{igl2018deep, ha2018world, zintgrafvariational}.  We first propose a VAE for learning opponent representations in multi-agent systems based on the opponent trajectories. 


A shortcoming of this approach and most opponent modeling methods, as will be discussed in Section \ref{sec:rel_work}, is that they require access to opponent's information, such as observations and actions, during training as well as execution. This assumption is too limiting in the majority of scenarios. For example, consider Poker, in which agents do not have access to the opponent's observations. Nevertheless, during Poker, humans can reason about the opponent's behaviors and goals using only their local observations. For example, an increase in the table's pot could mean that the opponent either holds strong cards or is bluffing. Based on the idea that an agent can reason about an opponent models using its own observations, actions, and rewards in a recurrent fashion, we propose a second VAE-based architecture. The encoder of the VAE learns to represent opponent models conditioned on only local information removing the requirement to access the opponents' information during execution. 

We evaluate our proposed methodology using a toy example and the commonly used Multi-agent Particle Environment \citep{mordatch2017emergence}. We evaluate the episodic returns that RL algorithms can achieve. The experiments indicate that opponent modeling without opponents' information can achieve comparable or higher average returns, using RL, than models that access the opponent's information.

\section{Related Work}
\label{sec:rel_work}

\textbf{Learning Opponent Models.} In this work, we are interested in opponent modeling methods that use neural networks to learn representations of the opponents. \citet{he2016opponent} proposed an opponent modeling method that learns a modeling network to reconstruct the opponent's actions given the opponent observations. \citet{raileanu2018modeling} developed an algorithm for learning to infer opponents' goals using the policy of the controlled agent. \citet{hernandez2019agent} used auxiliary tasks for modeling opponents in multi-agent reinforcement learning. \citet{grover2018learning} proposed an encoder-decoder method for modeling the opponent's policy.  The encoder learns a point-based representation of different opponents' trajectories, and the decoder learns to reconstruct the opponent's policy given samples from the embedding space. Additionally, \citet{grover2018learning} introduce an objective to separate embeddings of different agents into different clusters. 
\begin{equation}
\label{eq:disc_obj}
    d(z_+, z_-, z) = \frac{1}{(1 + e^{|z - z_-|_2 - |z - z_+|_2})^2}
\end{equation}
where $z_+$ and $z$ are embeddings of the same agent from two different episodes and
embedding $z_-$ is generated from the episode of a different agent.
\citet{zheng2018deep} use an opponent modeling method for better opponent identification and multi-agent reinforcement learning. \citet{rabinowitz2018machine} proposed the Theory of mind Network (TomNet), which learns embedding-based representations of opponents for meta-learning. \citet{tacchetti2018relational} proposed \textit{Relation Forward Model} to model opponents using graph neural networks. A common assumption among these methods, which our work aims to eliminate, is that access to opponents trajectories is available during execution.

\textbf{Representation Learning in Reinforcement Learning.}
Another related topic that has received significant attention is representation learning in RL. Using unsupervised learning techniques to learn low-dimensional representations of the MDP has led to significant improvement in RL. \citet{ha2018world} proposed a VAE-based and a forward model to learn state representations of the environment.  \citet{hausman2018learning} learned tasks embeddings and interpolated them to solve harder tasks. \citet{igl2018deep} used a VAE for learning representation in partially-observable environments. \citet{gupta2018meta} proposed a model which learns Gaussian embeddings to represent different tasks during meta-training and manages to quickly adapt to new task during meta-testing. The work of \citet{zintgrafvariational} is closely related, where \citeauthor{zintgrafvariational} proposed a recurrent VAE model, which receives as input the observation, action, reward of the agent, and learns a variational distribution of tasks. \citet{rakelly2019efficient} used representations from an encoder for off-policy meta-RL. Note that all these works have been applied for learning representations of tasks or properties of the environments. In contrast, our approach is focused on learning representations of opponents.


\section{Background}
\label{sec:back}
\subsection{Reinforcement Learning} 
Markov Decision Processes (MDPs) are commonly used to model decision making problems. An MDP consists of the set of states $\sS$, the set of actions $\sA$, the transition function, $P(\vs'| \vs,\va)$, which is the probability of the next state, $s'$, given the current state, $s$, and the action, $a$, and the reward function, $r(\vs',\va,\vs)$, that returns a scalar value conditioned on two consecutive states and the intermediate action. A policy function is used to choose an action given a state, which can be stochastic $\va \sim \pi(\va | \vs)$ or deterministic $\va = \mu(\vs)$. Given a policy $\pi$, the state value function is defined as $V(\vs_t) = \E_{\pi} [\sum_{i=t}^{H}\gamma^{i-t} r_t |s=\vs_t] $ and the state-action value (Q-value) $Q(\vs_t, \va_t) = \E_{\pi} [\sum_{i=t}^{H}\gamma^{i-t} r_t |s=\vs_t, a=\va_t]$, where $0 \leq \gamma \leq 1$ is the discount factor and $H$ is the finite horizon of the episode. The goal of RL is to compute the policy that maximizes state value function $V$, when the transition and the reward functions are unknown.

There is a large number of RL algorithms; however, in this work, we focus on two actor-critic algorithms; the synchronous Advantage Actor-Critic (A2C) \citep{mnih2016asynchronous, baselines} and the Deep Deterministic Policy Gradient (DDPG) \citep{silver2014deterministic, lillicrap2015continuous}. DDPG is an off-policy algorithm, using an experience replay for breaking the correlation between consecutive samples and target networks for stabilizing the training \citep{mnih2015human}. Given an actor network with parameters $\vtheta$ and a critic network with parameter $\vphi$, the gradient updates are performed using the following update rules.
\begin{equation*}
\label{eq:ddpg_loss}
\begin{aligned}
&\min_{\vphi}\frac{1}{2}\E_{B}[ (r + \gamma \cdot Q_{\mathrm{target}, \vphi'}(\vs', \mu_{\mathrm{target},\vtheta'}(\vs')) - Q_{\vphi}(\vs, \va))^2] \\
& \min_{\vtheta} -\E_{B}[Q_{\phi}(\vs, \mu_\vtheta(\vs))]
\end{aligned}
\end{equation*}
On the other hand, A2C is an on-policy actor-critic algorithm, using parallel environments to break the correlation between consecutive samples. The actor-critic parameters are optimized by:

\begin{equation*}
\label{eq:a2c_loss}
\min_{\vtheta, \vphi}\E_{B}[- \hat{A}\log\pi_\theta(\va|\vs) + \frac{1}{2}(r + \gamma V_{\vphi}(\vs') - V_{\vphi}(\vs))^2 ]
\end{equation*}
where the advantage term, $\hat{A}$, can be computed using the Generalized Advantage Estimation (GAE) \citep{schulman2015high}. 
\subsection{Variational Autoencoders}
Consider samples from a dataset $\vx\in \sX$ which are generated from some hidden (latent) random variable  $\vz$ based on a generative distribution $p_{\vu}(\vx|\vz)$ with unknown parameter $\vu$ and a prior distribution on the latent variables, which we assume is a Gaussian with $\vzero$ mean and unit variance $p(\vz)  = \mathcal{N}(\vz; \vzero , \vI) $. We are interested in approximating the true posterior $p(\vz|\vx)$ with a variational parametric distribution $q_{\vw}(\vz | \vx) = \mathcal{N} ( \vz ; \vmu , \mSigma, \vw)$. 
\citet{kingma2013auto} proposed the Variational Autoencoders (VAE) to learn this distribution. Starting from the Kullback-Leibler (KL) divergence from the approximate to the true posterior $\KL ( q_{\vw}(\vz| \vx) \Vert p(\vz|\vx) )$, the lower bound on the evidence $\log p(\vx)$ is derived as:
\begin{equation*} \log p(\vx) \geq \E_{\rz\sim q_{\vw}(\vz| \vx)}[\log p_{\vu}(\vx | \vz)]  - \KL( q_{\vw}(\vz| \vx) \Vert p(\vz))
\end{equation*}
The architecture consists of an encoder which receives a sample $\vx$ and generates the Gaussian variational distribution $p(\vz| \vx; \vw)$. The decoder receives a sample from the Gaussian variational distribution and reconstructs the initial input $\vx$. The architecture is trained using the reparameterization trick \citet{kingma2013auto}.
\citet{higgins2017beta} proposed $\beta$-VAE, where a parameter $\beta \geq 0$ is used to control the trade-off between the reconstruction loss and the KL-divergence.
\begin{equation*}
\begin{aligned}
& L(\vx; \vw, \vv) =  \E_{\rz\sim q_{\vw}(\vz| \vx)}[\log p_{\vu}(\vx | \vz)]  \\
& - \beta\KL( q_{\vw}(\vz| \vx) \Vert p(\vz))
\end{aligned}
\end{equation*}

\section{Approach}
\label{sec:method}

\subsection{Problem Formulation}

We consider a modified Markov Game \citep{littman1994markov}, which consists of $N$ agents $\sI = \{1,2,...,N\}$, a set of states $\sS$, the joint action space $\sA = \sA_1 \times \sA_{-1}$ , the transition function $P:\sS\times \sA \times \sS \rightarrow \mathbb{R}$ and the reward function $r:\sS\times \sA \times \sS \rightarrow \mathbb{R}^N$. We consider partially observable settings, where each agent $i$ has access only to its local observation $\vo_i$ and reward $r_i$. Additionally, two sets of pretrained opponents are provided, $\sT = \{\sI_{-1, m}\}_{m=1}^{m=M_T}$ and $\sG = \{\sI_{-1, m}\}_{m=1}^{m=M_G}$, which are responsible for providing the joint action $\sA_{-1}$. Note that by opponent we refer to $\sI_{-1,m}$, which consists of one or more agents, independently from the type of the interactions. Additionally, we assume that the sets $\sT$ and $\sG$ are disjoint.
At the beginning of each episode, we sample a pretrained opponent from the set $\sT$ during training or from $\sG$ during testing. Our goal is to train agent $1$ using RL, to maximize the average return against opponents from the training set, $\sT$, and generalize to opponents sampled from the test $\sG$. 
 \begin{equation}
     \arg \max_{\theta}   \E_{\pi}[\E_{\sT}[\sum_{t} \gamma^t r_{t}]]
 \end{equation}
Note, that when we refer to agent $1$ we drop the subscript. 

\subsection{Variational Autoencoder with Access to Opponent's Information}
\label{sec:vae_opp_info}
\begin{figure}[H]
    \centering
    \includegraphics[width=\linewidth]{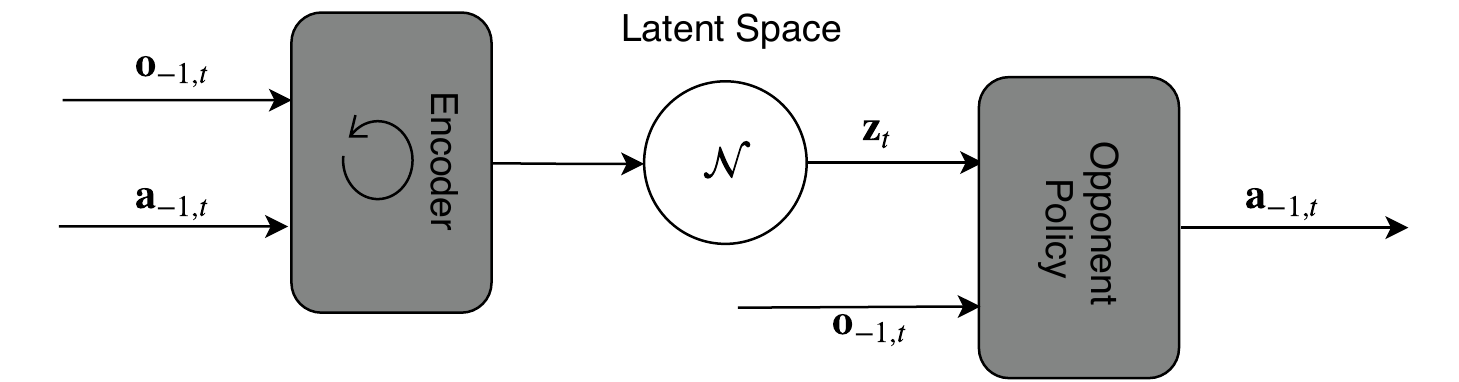}
    \caption{Diagram of the proposed VAE architecture}
    \label{fig:pre_vae}
\end{figure}

We assume a number of $K$ provided episode trajectories for each pretrained opponent $j \in \sT $, $E^{(j)} = \{\vtau_{-1}^{(j, k)}\}_{k=0}^{k=K-1} $, where $\vtau_{-1}^{(j, k)} = \{\vo_{-1,t}, \va_{-1,t} \}_{t=0}^{t=H}$, and $\vo_{-1,t}, \va_{-1,t}$ are the observations and actions of the opponent at the time step $t$ in the trajectory. These trajectories are generated from the opponents in set $\sT$, which are represented in the latent space from the variable $\vz$ and for which we assume there exists an unknown model $p_{\vu}(\vtau_{-1}|\vz)$. Our goal is to approximate the unknown posterior, $p(\vz|\vtau_{-1})$, using a variational Gaussian distribution $\mathcal{N}(\vmu, \mSigma ; \vw)$ with parameters $\vw$. We consider using a $\beta$-VAE for the sequential task:
\begin{equation*}
\begin{aligned}
& L(\vtau_{-1}; \vw, \vu) =  \E_{\rz\sim q_{\vw}(\vz| \vtau_{-1})}[\log p_{\vu}(\vtau_{-1} | \vz)]  \\
& - \beta\KL( q_{\vw}(\vz| \vtau_{-1}) \Vert p(\vz))
\end{aligned}
\end{equation*}

We can further subtract the discrimination objective (equation \ref{eq:disc_obj}) that was proposed by \cite{grover2018learning}. Since the discrimination objective is always non-negative, we derive and optimize a lower bound  as:
\begin{equation*}
\label{eq:opp_vae_loss}
\begin{aligned}
  & L(\vtau_{-1}; \vw, \vu) \geq \E_{\rz\sim q_{\vw}(\vz| \vtau_{-1})}[\log p_{\vu}(\vtau_{-1} | \vz)] \\
  & - \beta\KL( q_{\vw}(\vz| \vtau_{-1}) \Vert p(\vz)) - \lambda d(\E(\vz_{+}), \E(\vz_{-}), \E(\vz))  
 \end{aligned}
\end{equation*}
The discrimination objective receives as input the mean of the variational Gaussian distribution, produced by three different trajectories. Despite the less tight lower bound, the discrimination objective will separate the opponents in the embedding space, which could potentially lead to higher episodic returns. At each time step $t$, the recurrent encoder network generates a latent sample $z_t$, which is conditioned on the opponent's trajectory $\vtau_{-1, :t}$, until time step $t$. The KL divergence can be written as:
\begin{equation*}
    \KL( q_{\vw}(\vz| \vtau_{-1}) \Vert p(\vz)) = \sum_{t=1}^{H} \KL( q_{\vw}(\vz_t| \vtau_{-1, :t}) \Vert p(\vz_t))
\end{equation*}
The lower bound consist of the reconstruction loss of the trajectory which involves the observation and actions of the opponent. The opponent's action, at each time step depends on its observation and the opponent's policy, which is represented by the latent variable $z$. We use a decoder that consists of fully-connected layers, however, a recurrent  network can be used, if we instead assume that the opponent decides its actions based on the history of its observations. Additionally, the observation at each time step depends only on the dynamics of the environment and the actions of the agents and not on the identity of the opponent. Therefore, the reconstruction loss factorizes as:
\begin{equation}
\label{recon_vae}
\begin{aligned}
& \log p_{\vu}(\vtau_{-1}| \vz) =  \sum_{t=1}^{H}  \log  p_{\vu}(\va_{-1,t} |\vo_{-1,t}, \vz_t)  \\
& p_{\vu}(\vo_{-1,t} |\vs_{t-1}, \va_{t-1}, \va_{-1, t-1}) \\
&\propto \sum_{t=1}^{H}  \log  p_{\vu}(\va_{-1,t} |\vo_{-1,t}, \vz_t)
\end{aligned}
\end{equation}
From the equation above, we observe that the loss is the reconstruction of the opponent's policy  given the current observation and a sample from the latent variable. Figure \ref{fig:pre_vae} illustrates the diagram of the VAE. 

\label{sec:toy_example}
\begin{figure*}
    \centering 
\begin{subfigure}{0.33\textwidth}
\centering
  \includegraphics[width=0.55\linewidth]{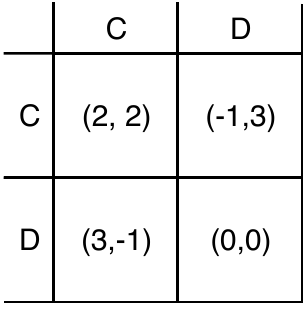}
  \caption{Payoff matrix}
  \label{fig:SMA2C_emb}
\end{subfigure}\hfil 
\begin{subfigure}{0.33\textwidth}
  \includegraphics[width=\linewidth]{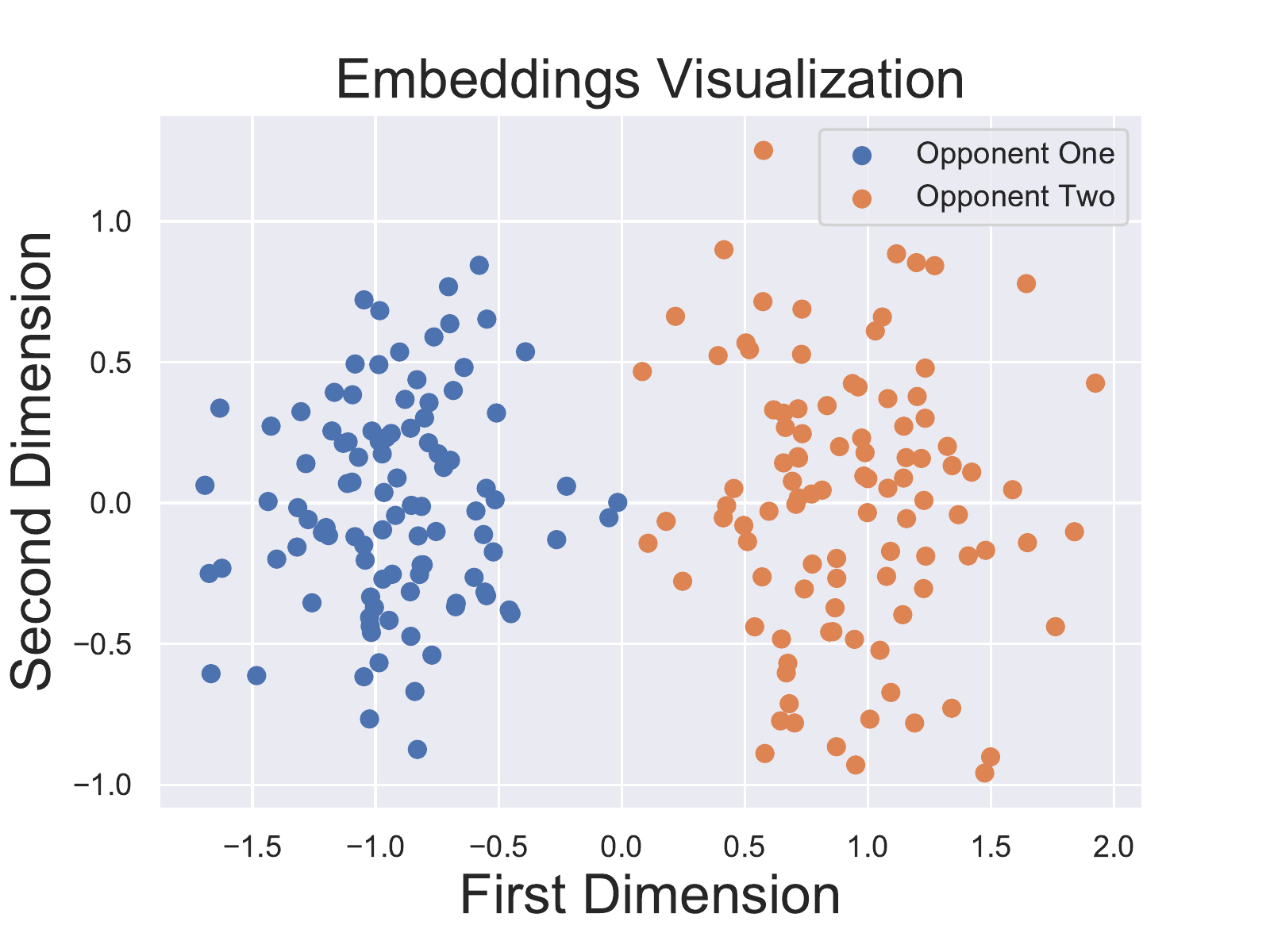}
  \caption{Embedding visualization}
  \label{fig:omddpg_emb}
\end{subfigure}\hfil 
\begin{subfigure}{0.33\textwidth}
\includegraphics[width=\linewidth]{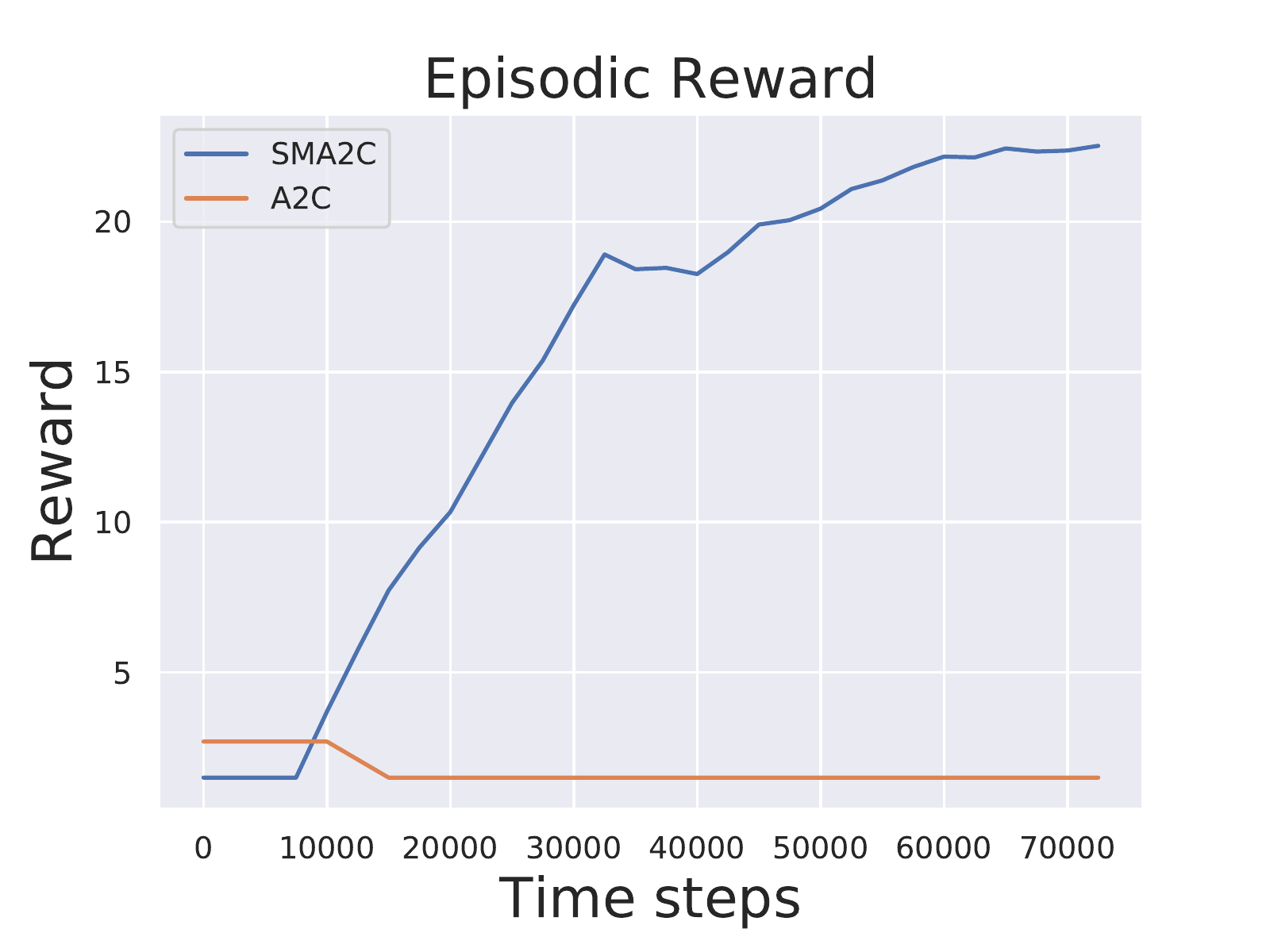}
  \caption{Episodic return}
  \label{fig:omddpg_no_dicr_emb}
\end{subfigure}\hfil %
\caption{Payoff matrix, embedding visualization and episodic return during training in the toy example.}
\label{fig:toy_example}
\end{figure*}

\subsection{Variational Autoencoder without Access to Opponent's Information}

\begin{figure}[H]
    \centering
    \includegraphics[width=\linewidth]{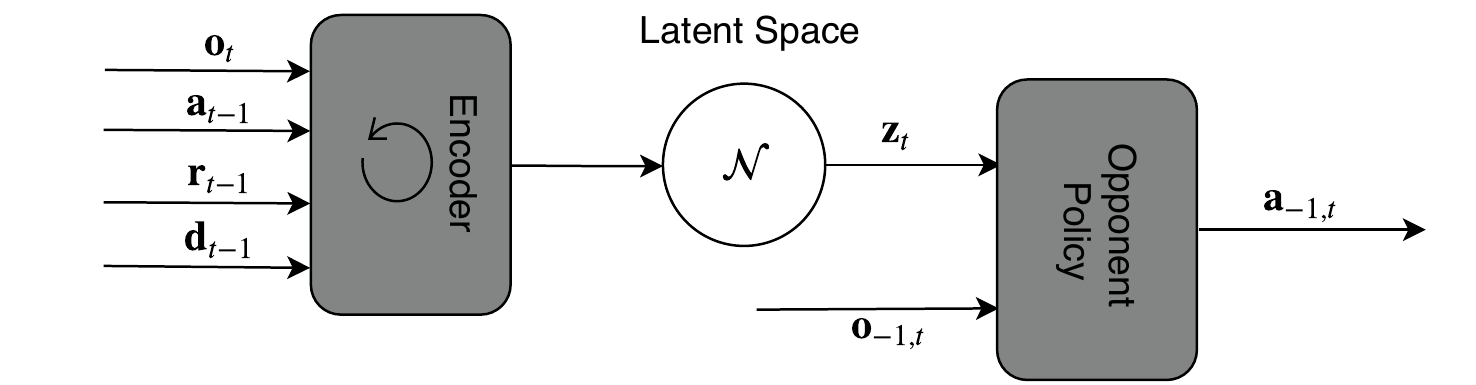}
    \caption{Diagram of the proposed VAE architecture using local information}
    \label{fig:self_vae}
\end{figure}
In Sections \ref{sec:intro} and \ref{sec:rel_work}, it was noted that most agent modeling methods assume access to opponent's observations and actions is available both during  training and execution. To eliminate this assumption, we propose a VAE that uses a parametric variational distribution which is conditioned on the observation-action-reward triplet of the agent under our control and a variable $d$ indicating whether the episode has terminated; $q_{\vw}(\vz | \vtau=(\vo,\va,r,d))$. Specifically, our goal is to approximate the true posterior that is conditioned on opponent's information, with a variational distribution that only depends on local information. The use of this local information in a recurrent fashion has been successfully used in meta-RL settings \citep{wang2016learning,duan2016rl}. We start by computing the KL divergence between the two distributions:
\begin{equation*}
\begin{aligned}
&  \KL ( q_{\vw}(\vz| \vtau) \Vert p(\vz|\vtau_{-1}) ) = \E_{\rz\sim q_{\vw}(\vz| \vtau)}[\log q_{\vw}(\vz| \vtau) \\
&- \log p(\vz|\vtau_{-1})]
\end{aligned}
\end{equation*}
By following the works of \citet{kingma2013auto} and \citet{higgins2017beta} and using the Jensen inequality, the VAE objective can be written as:
\begin{equation}
\label{eq:self_vae}
\begin{aligned}
&  L(\vtau, \vtau_{-1}; \vw, \vv) =  \E_{\rz\sim q_{\vw}(\vz| \vtau)}[\log p_{\vu}(\vtau_{-1} | \vz)]  \\
& - \beta\KL( q_{\vw}(\vz| \vtau) \Vert p(\vz))
\end{aligned}
\end{equation}

The reconstruction loss factorizes exactly similar to equation \ref{recon_vae}. From equation \ref{eq:self_vae}, it can be seen that the variational distribution only depends on locally available information. Since during execution only the encoder is required to generate the opponent's model, this approach removes the assumption that access to the opponent's observations and actions is available during execution. Figure \ref{fig:self_vae} presents the diagram of the VAE.

\subsection{Reinforcement Learning Training}

We use the latent variable $\vz$ augmented with the agent's observation to condition the policy of our agent, which is optimized using RL. Consider the augmented observation space $\sO' = \sO \times \sZ$, where $\sO$ is the original observation space of our agent in the Markov game, and $\sZ$ is the representation space of the opponent models. The advantage of learning the policy on $\sO' $ compared to $\sO$ is that the policy can adapt to different $\vz \in \sZ$.

After training the variational autoencoder described in Section \ref{sec:vae_opp_info}, we use it to train our agent against the opponents in the set $\sT$. We use the DDPG \citep{lillicrap2015continuous} algorithm for this task. At the beginning of each episode, we sample an opponent from the set $\sT$. The agent's input is the local observation and a sample from the variational distribution.  We refer to this as OMDDPG (Opponent Modeling DDPG).


We optimize the second proposed VAE method jointly with the policy of the controlled agent. We use the A2C algorithm, similarly to the meta-learning algorithm  RL$^2$ \citep{wang2016learning,duan2016rl}. In the rest of this paper, we refer to this as SMA2C (Self Modeling A2C). The actor's and the critic's input is the local observation and a sample from the latent space. We back-propagate the gradient from both the actor and the critic loss to the parameters of the encoder. Therefore, the encoder's parameters are shaped to maximize both the VAE's objective as well as the discounted sum of rewards. 

\section{Experiments}

\subsection{Toy Example}
We will first provide a toy example to illustrate SMA2C. We consider the classic repeated game of prisoner's dilemma with a constant episode length of $25$ time steps. We control one agent, and the opponent uses one of 
two possible policies. The first policy always defects, while the second opponent policy follows a tit-for-tat policy. At the beginning of the episode, one of the two opponent policies is randomly selected. We train SMA2C against the two possible opponents. The agent that we control has to identify the correct opponent, and the optimal policy, it can achieve, is to defect against opponent one and collaborate with opponent two. Figure \ref{fig:toy_example} shows the payoff matrix, the embedding space at the last time step of the episode, and the episodic return that SMA2C and A2C achieve during training. Note that, based on the payoff matrix, the optimal average episodic return that can be achieved is $24.5$. 

\begin{figure*}
    \centering 
\begin{subfigure}{0.245\textwidth}
  \includegraphics[width=\linewidth]{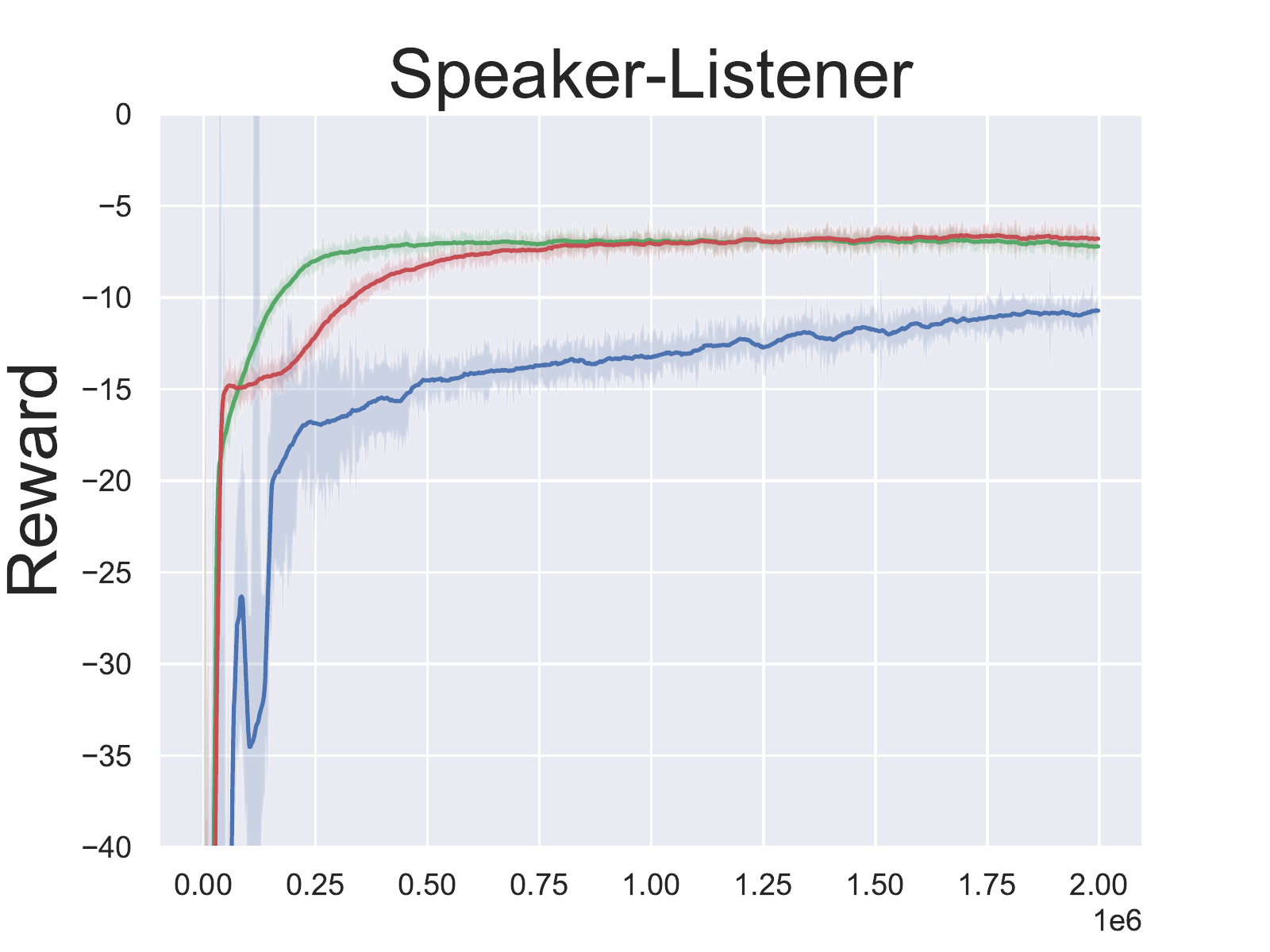}
  \label{fig:speaker_listener_wrl}
\end{subfigure}
\begin{subfigure}{0.245\textwidth}
  \includegraphics[width=\linewidth]{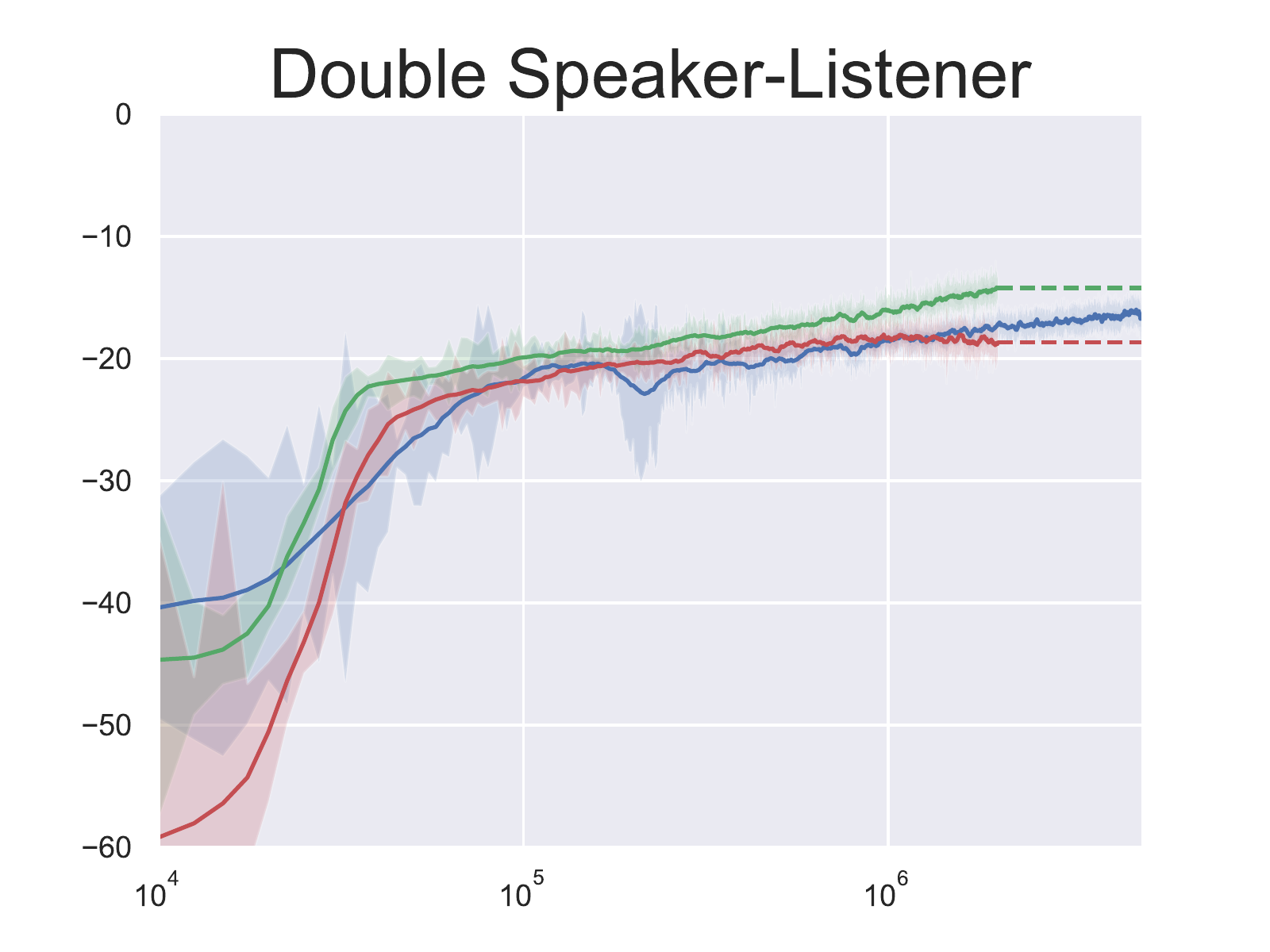}
  \label{fig:double_speaker_listener_wrl}
\end{subfigure}\hfil 
\begin{subfigure}{0.245\textwidth}
 \includegraphics[width=\linewidth]{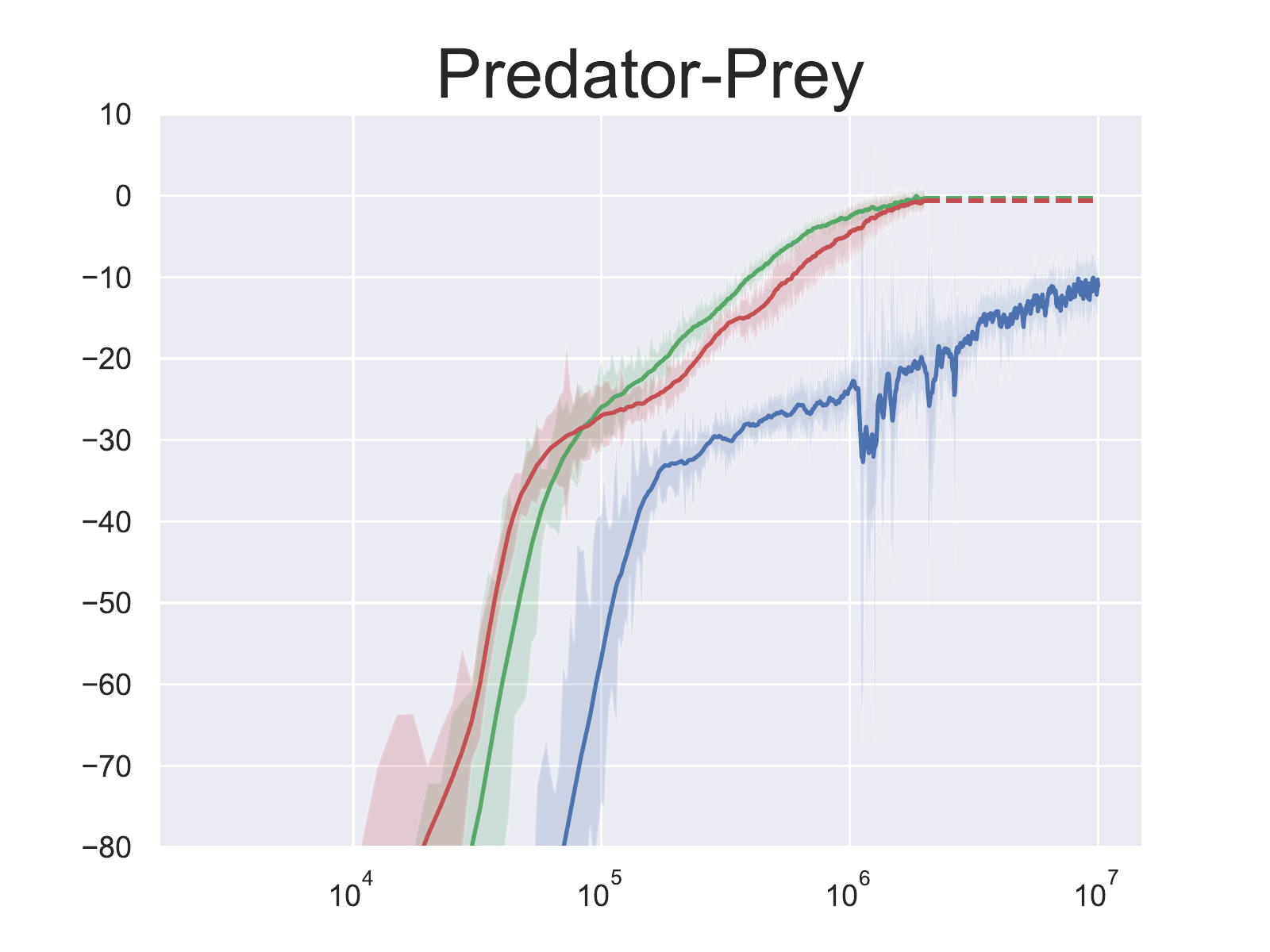}
  \label{fig:predator_prey_wrl}
\end{subfigure}\hfil 
\begin{subfigure}{0.245\textwidth}
  \includegraphics[width=\linewidth]{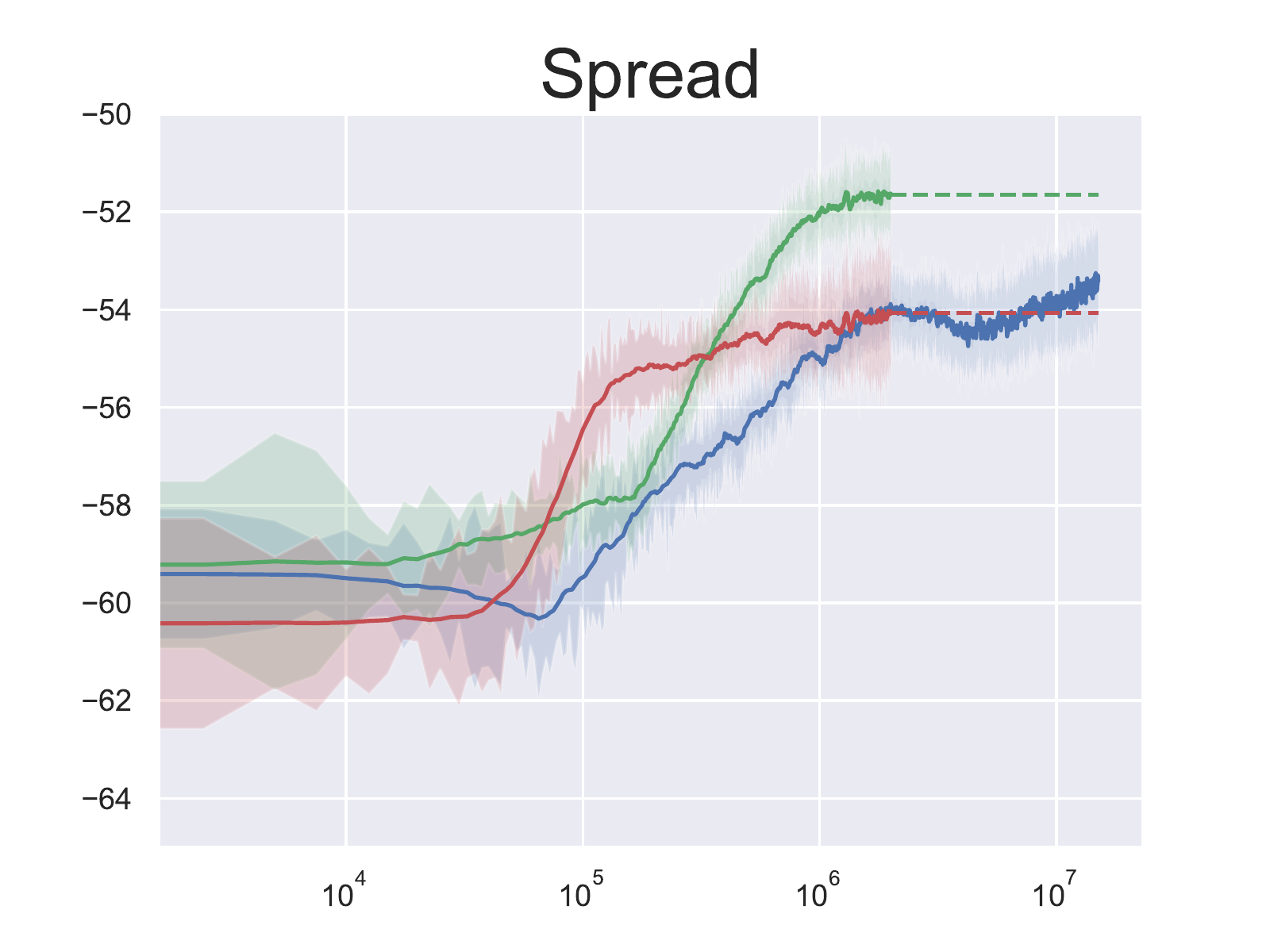}
  \label{fig:spread_wrl}
\end{subfigure} 
\centering
\begin{subfigure}{0.245\textwidth}
  \includegraphics[width=\linewidth]{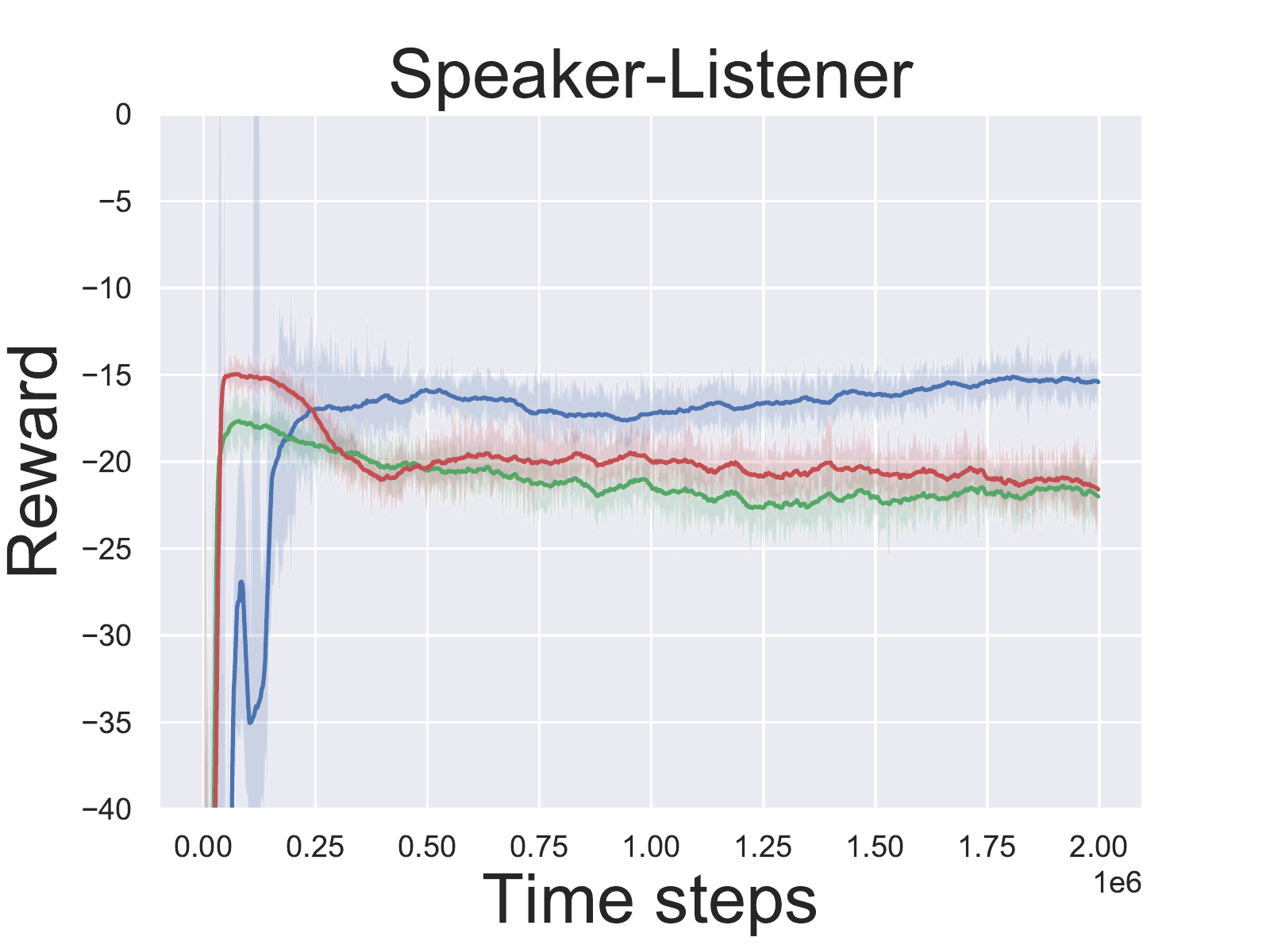}
  \label{fig:speaker_listener_srl}
\end{subfigure}
\begin{subfigure}{0.245\textwidth}
  \includegraphics[width=\linewidth]{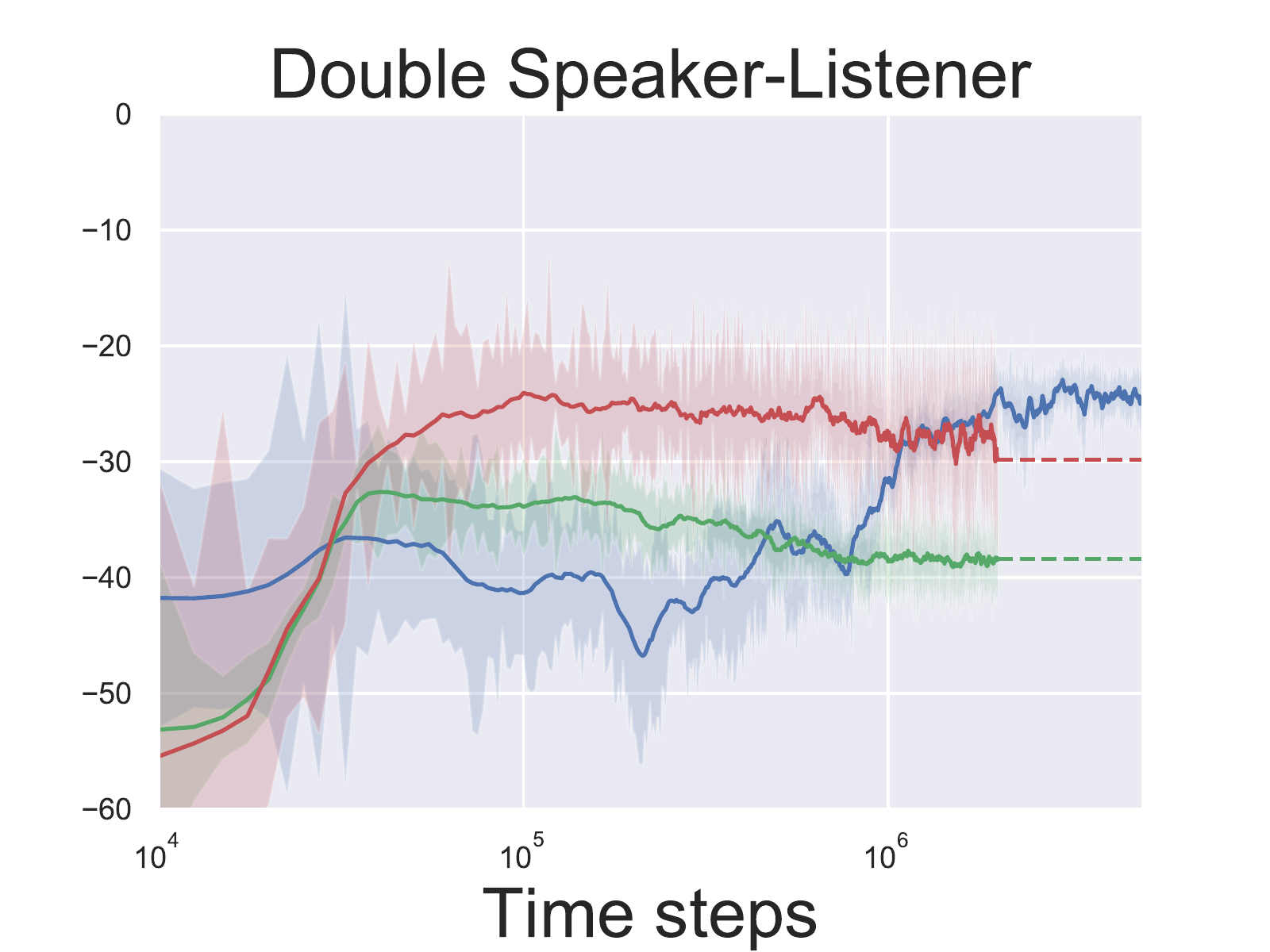}
  \label{fig:double_speaker_listener_srl}
\end{subfigure}
\begin{subfigure}{0.245\textwidth}
 \includegraphics[width=\linewidth]{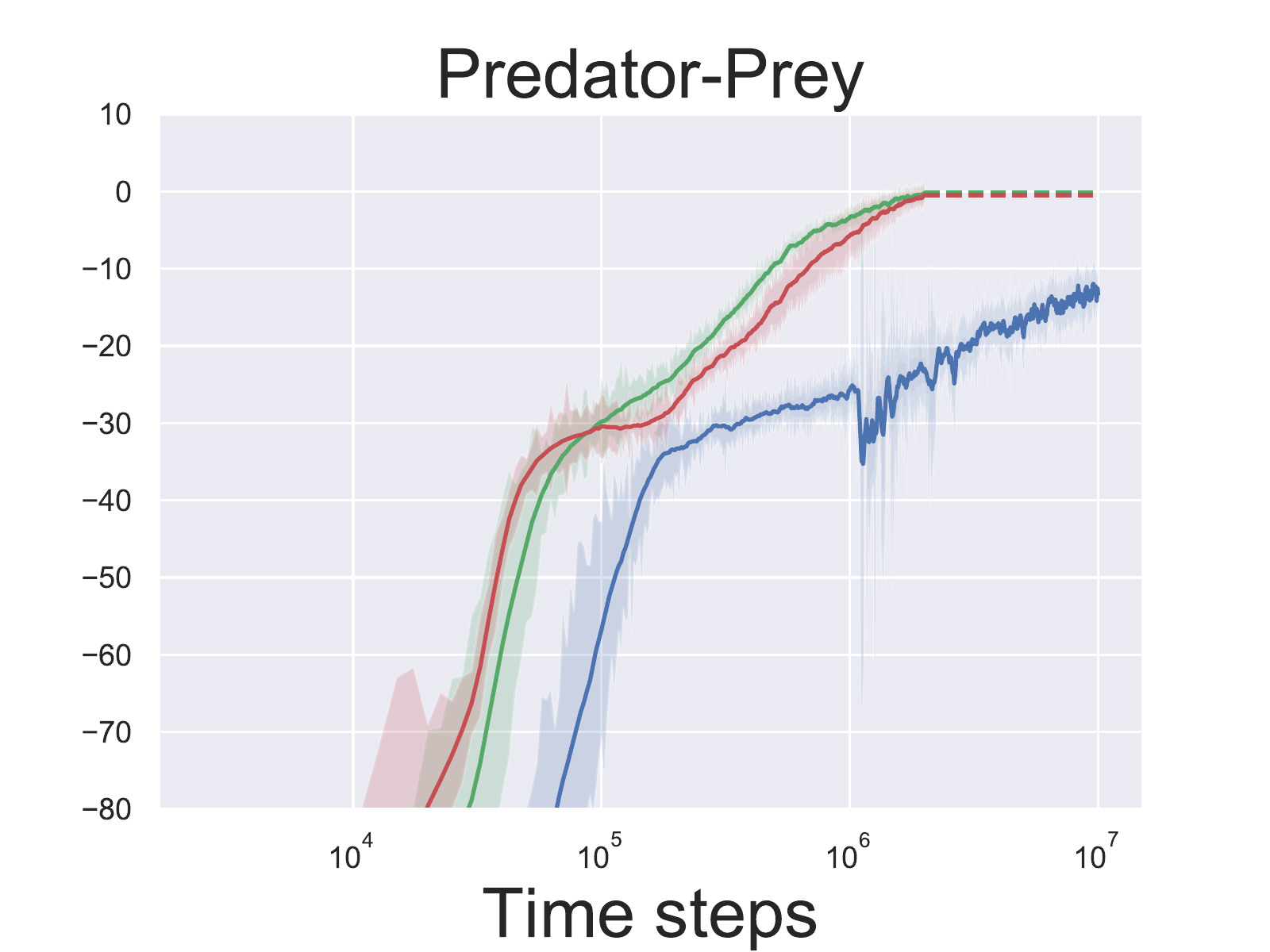}
  \label{fig:predator_prey_srl}
\end{subfigure}
\begin{subfigure}{0.245\textwidth}
  \includegraphics[width=\linewidth]{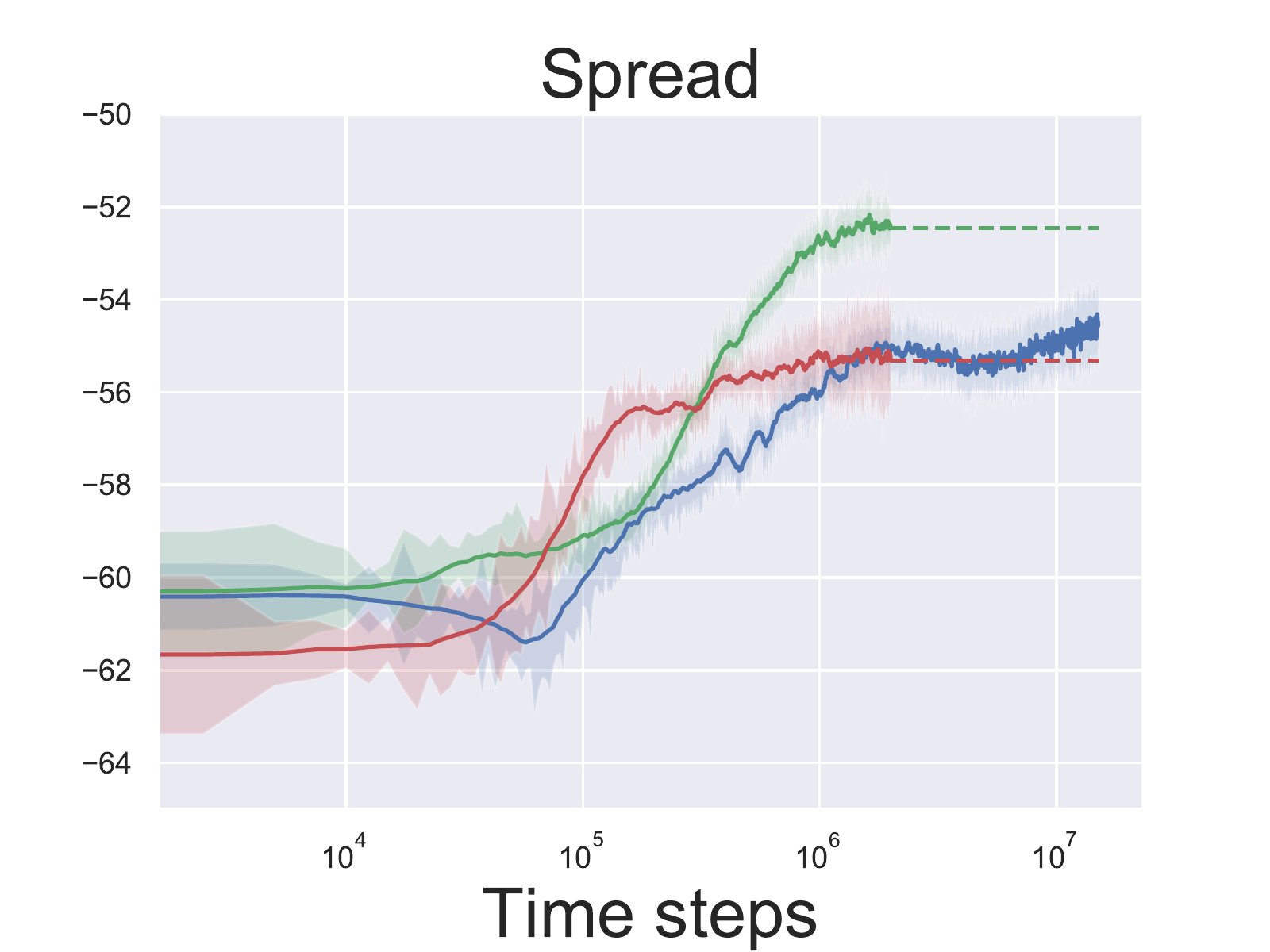}
  \label{fig:spread_srl}
\end{subfigure}
\fcolorbox{background_sns}{background_sns}
{ \smac \hspace{0.1cm}  SMA2C \hspace{1cm} \omddpg \hspace{0.1cm} OMDDPG \hspace{1cm} \grover \hspace{0.1cm} \citet{grover2018learning}
}
\caption{Episodic return during training against the opponents from $\sT$ (top row) and $\sG$ (bottom row). Four environments are evaluated; speaker-listener (first column), double speaker-listener (second column), predator-prey (third column) and spread (fourth column). For the double speaker-listener, predator-prey and spread environments the x-axis is in logarithmic scale.}
\label{fig:rl_perf}
\end{figure*}

\subsection{Experimental Framework}

To evaluate the proposed methods in more complex environments, we used the Multi-agent Particle Environment (MPE) \citep{mordatch2017emergence}, which provides several different multi-agent environments. The environments have continuous observation, discrete action space, and fixed-length episodes of $25$ time steps. Four environments are used for evaluating the proposed methodology; speaker-listener, double-speaker listener, predator-prey, and spread. During the experiments, we evaluated the two proposed algorithms OMDDPG and SMA2C and the modeling method of \citet{grover2018learning} combined with DDPG \citep{lillicrap2015continuous}.

In all the environments, we pretrain ten different opponents, where five are used for training and five for testing.  In the speaker-listener environment, we control the listener, and we create ten different speakers using different communication messages for different colors. In the double speaker-listener, which consists of two agents that have to be both listener and speaker simultaneously, we control the first agent. We create a diverse set of opponents that have different communication messages similar to speaker-listener, while they learn to navigate using the MADDPG algorithm \citep{lowe2017multi}, with different initial random seeds. In the predator-prey environment, we control the prey and pretrain the three other agents in the environment using MADDPG with different initial parameters. Similarly, in spread, we control one of the agents, while the opponents are pretrained using MADDPG.

We use agent generalization graphs \citep{grover2018evaluating} to evaluate the generalization of the proposed methods. We evaluate two types of generalizations. First, we evaluate the episodic returns against the opponents that are used for training, $\sT$, which \citet{grover2018evaluating} call "weak generalization". Secondly, we evaluate against unknown opponents from the set $\sG$, which is called "strong generalization".

\subsection{Reinforcement Learning Performance}

\begin{table*}
  \centering
    \caption{MI estimations using MINE in the double speaker-listener and the  predator-prey  environment of the embeddings at the $15$th, $20$th and  $25$th time step of the trajectory.}
  \begin{tabular}{|l | c c c || c c c |}
  \cline{2-7} 
   \multicolumn{1}{c|}{} & \multicolumn{3}{|c||}{Double speaker-listener}   & \multicolumn{3}{|c|}{Predator-prey} \\ \hline 
     \textbf{Algorithm\textbackslash  Time step} & $15$ & $20$ & $25$ & $15$ & $20$ & $25$ \\ \hline
    OMDDPG & $0.86$ & $0.81$  & $0.86$ & $0.87$ & $0.86$  & $0.85$ \\  \hline
    SMA2C & $0.73$ & $0.72$  & $0.7$ & $1.27$ & $1.04$  & $ 1.39$ \\  \hline
    \citet{grover2018learning} & $1.16$ & $1.21$ & $1.20$ & $1.29$ & $1.38$  & $1.41$\\  \hline
  \end{tabular}
  \label{tab:mine_valus}
\end{table*}

\begin{figure*}
    \centering 
\begin{subfigure}{0.25\textwidth}
  \includegraphics[width=\linewidth]{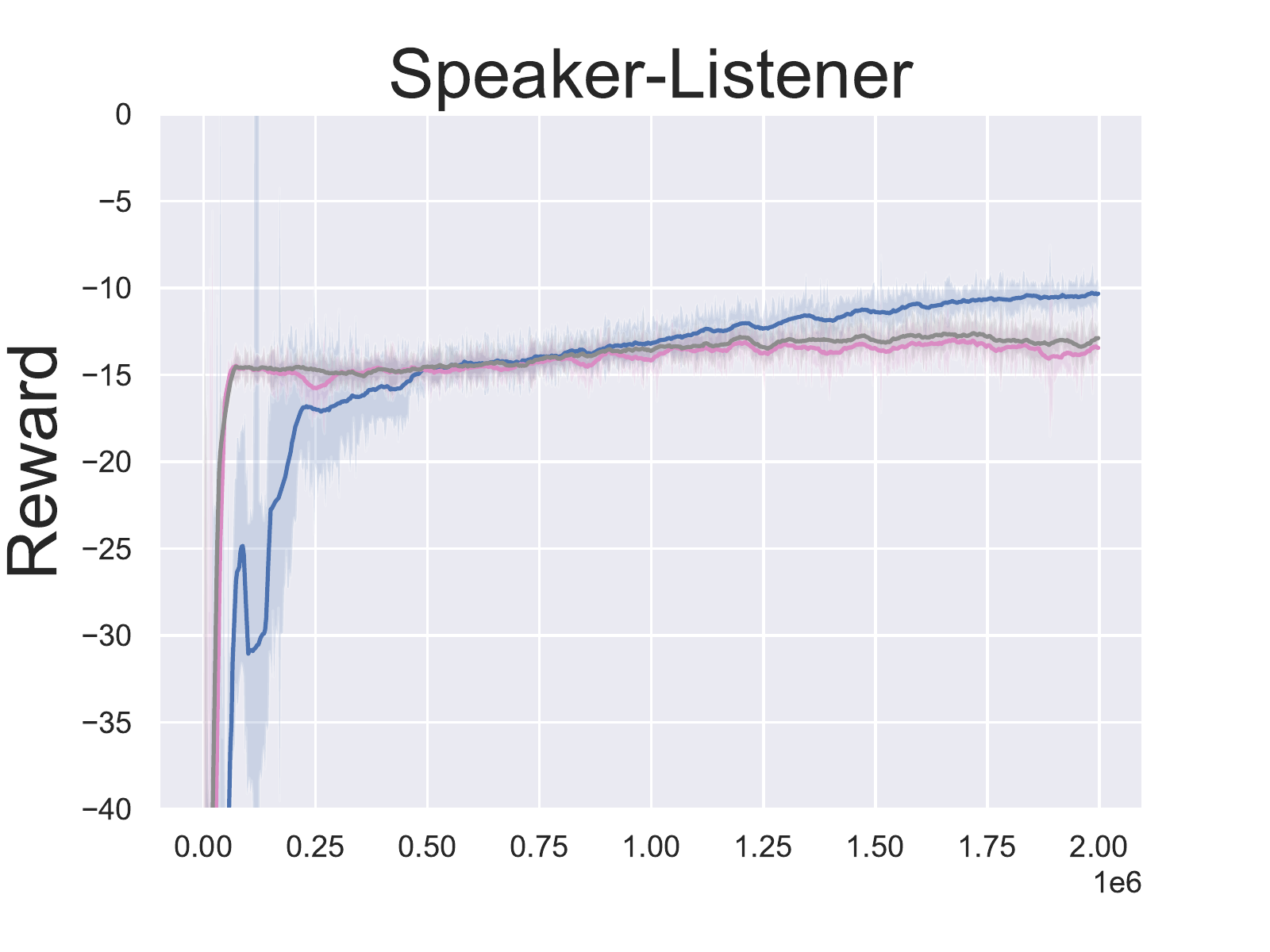}
  \label{fig:speaker_listener_wrl_disc_abl}
\end{subfigure}\hfil 
\begin{subfigure}{0.25\textwidth}
  \includegraphics[width=\linewidth]{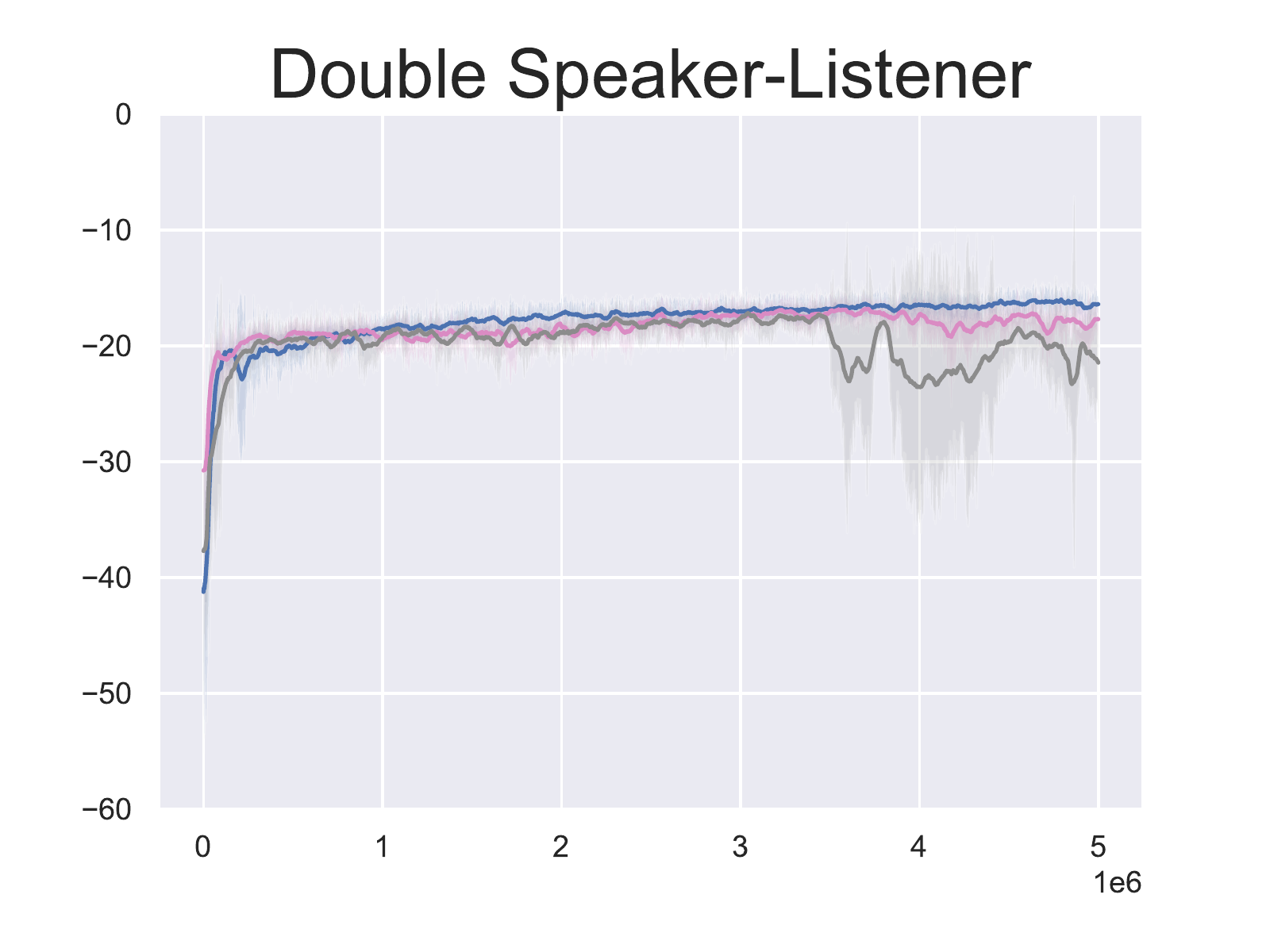}
  \label{fig:speaker_listener_srl_disc_abl}
\end{subfigure}\hfil 
\begin{subfigure}{0.25\textwidth}
 \includegraphics[width=\linewidth]{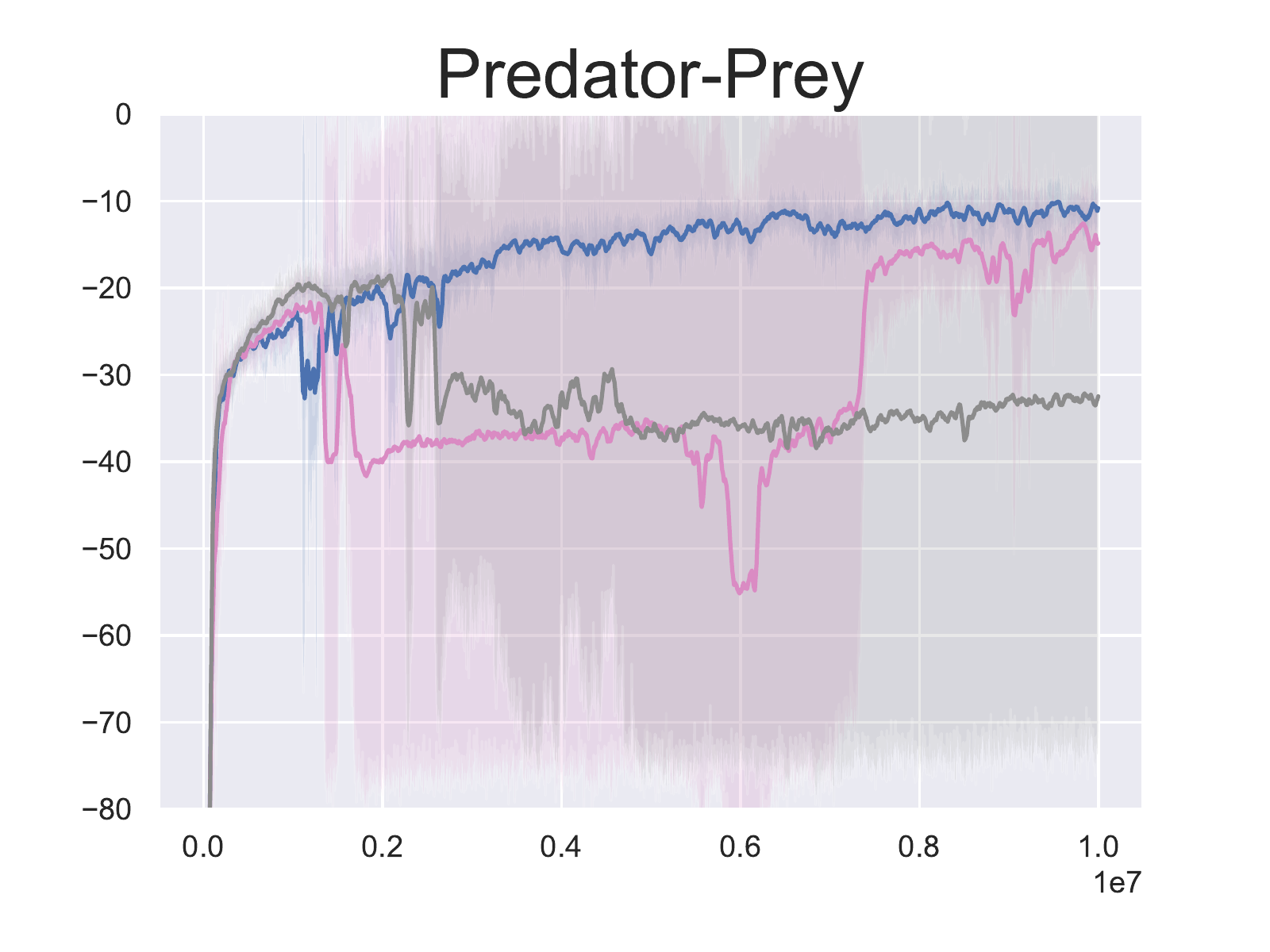}
  \label{fig:double_speaker_listener_wrl_disc_abl}
\end{subfigure}\hfil 
\begin{subfigure}{0.25\textwidth}
  \includegraphics[width=\linewidth]{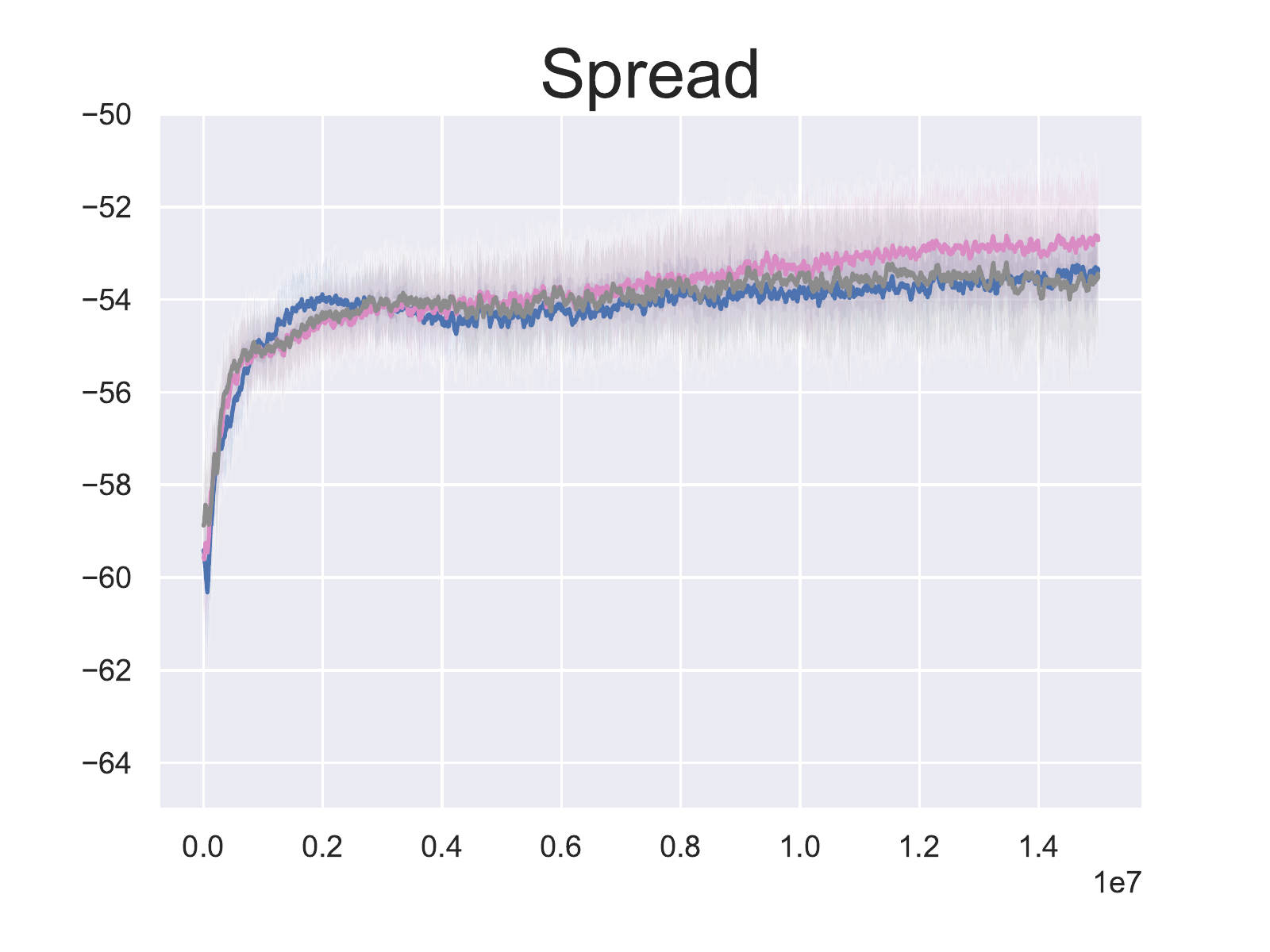}
  \label{fig:double_speaker_listener_srl_disc_abl}
\end{subfigure}\hfil 

    \centering 
\begin{subfigure}{0.25\textwidth}
  \includegraphics[width=\linewidth]{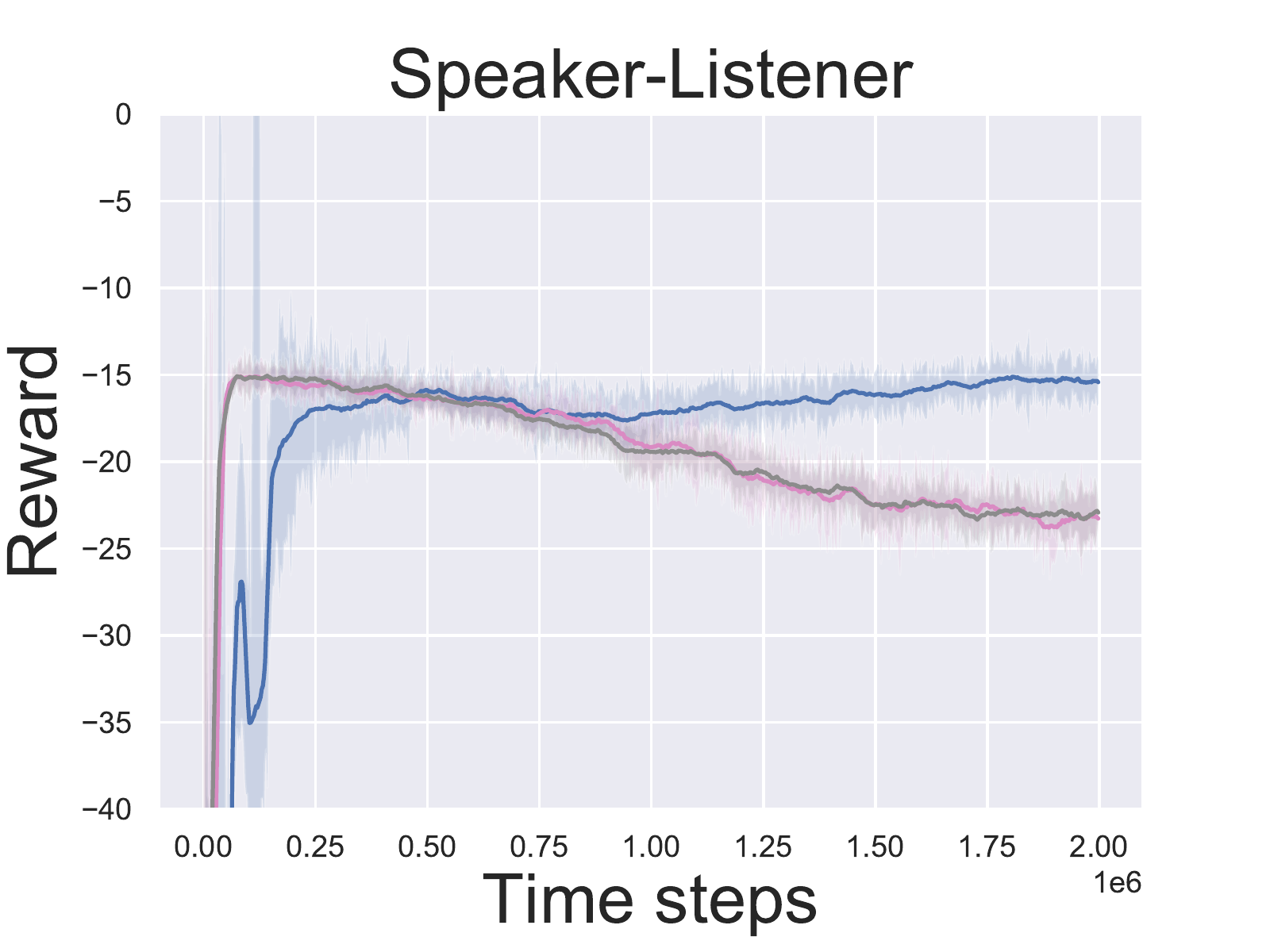}
  \label{fig:speaker_listener_wrl_disc_abl}
\end{subfigure}\hfil 
\begin{subfigure}{0.25\textwidth}
  \includegraphics[width=\linewidth]{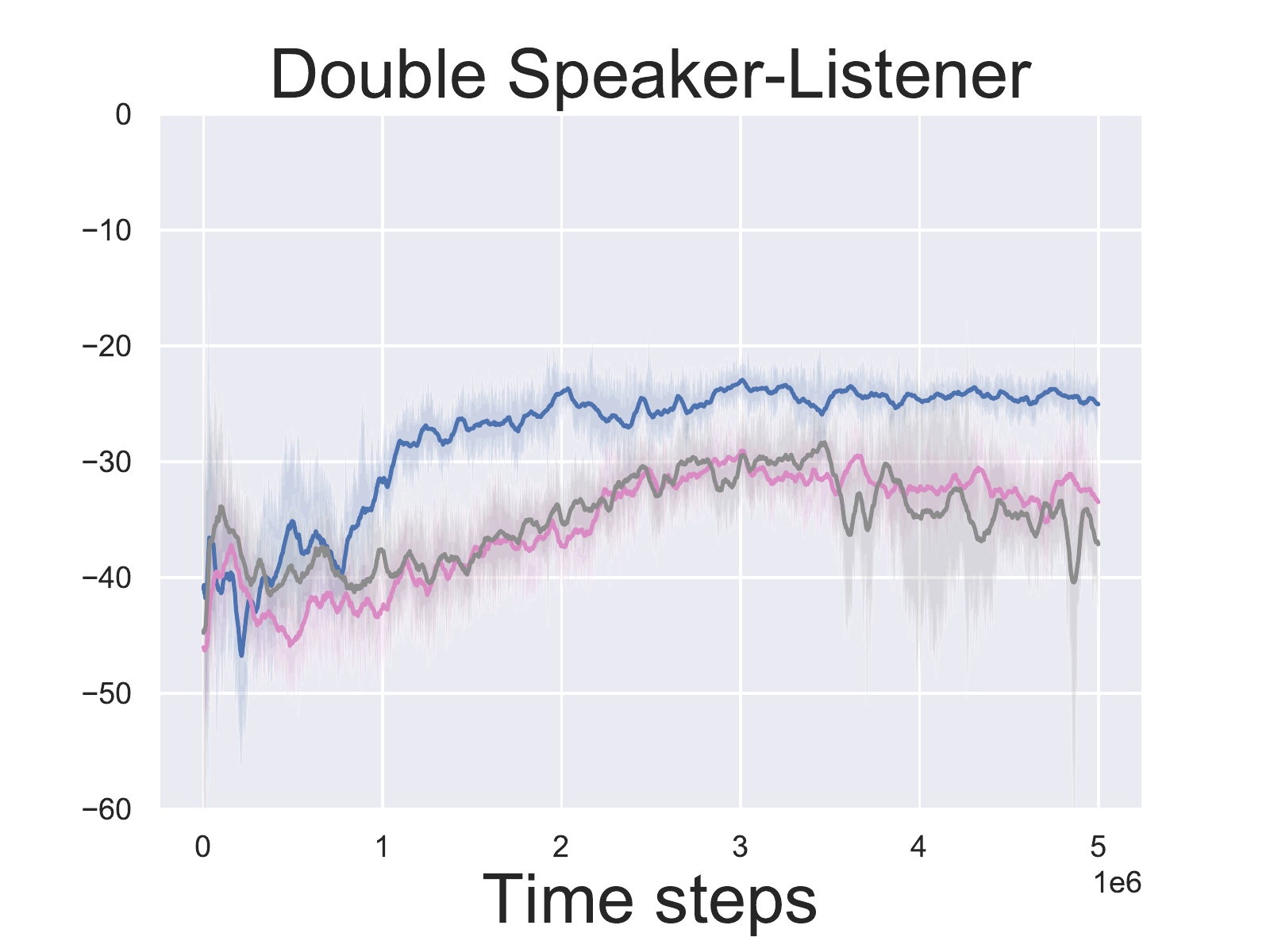}
  \label{fig:speaker_listener_srl_disc_abl}
\end{subfigure}\hfil 
\begin{subfigure}{0.25\textwidth}
 \includegraphics[width=\linewidth]{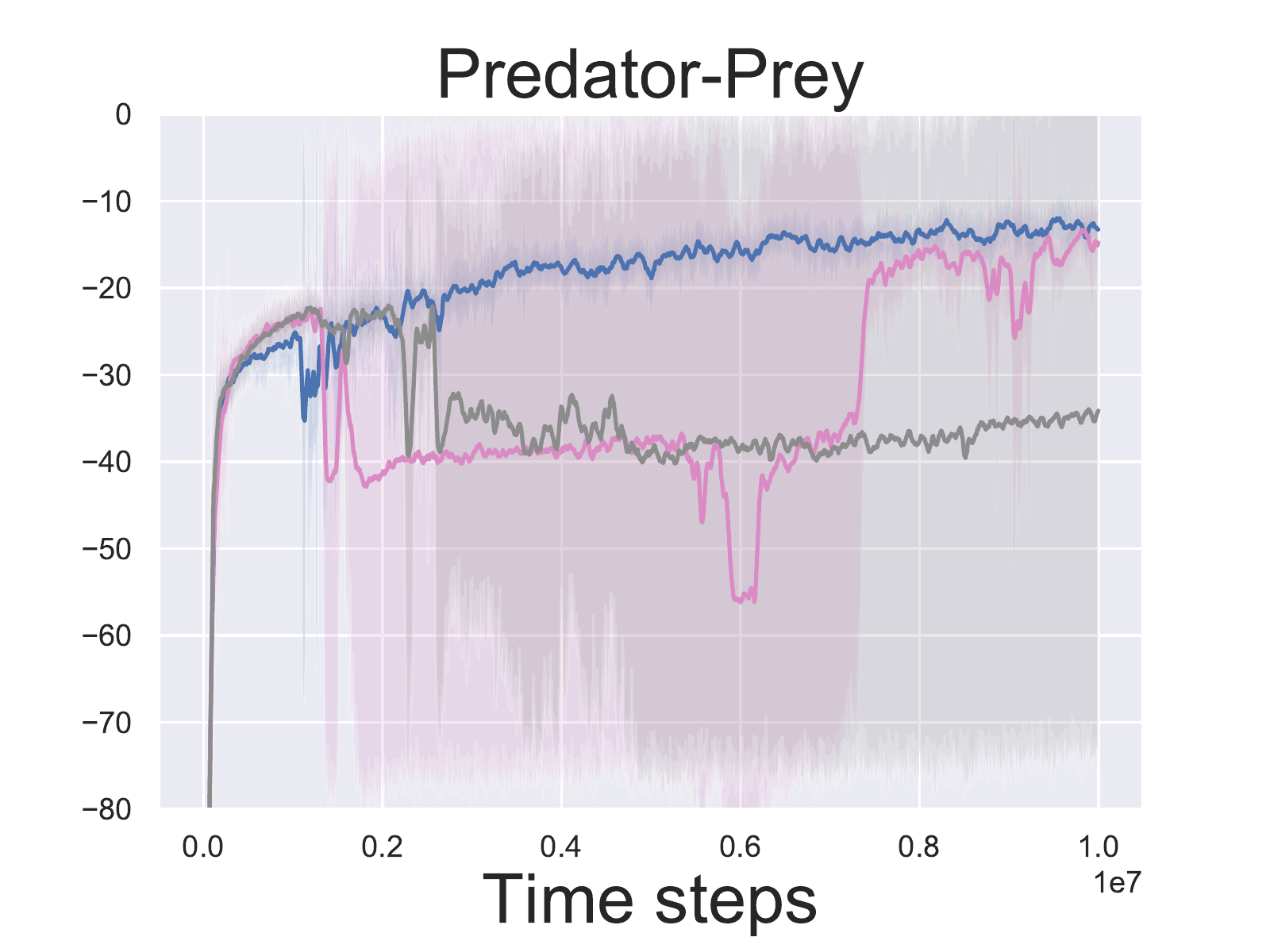}
  \label{fig:double_speaker_listener_wrl_disc_abl}
\end{subfigure}\hfil 
\begin{subfigure}{0.25\textwidth}
  \includegraphics[width=\linewidth]{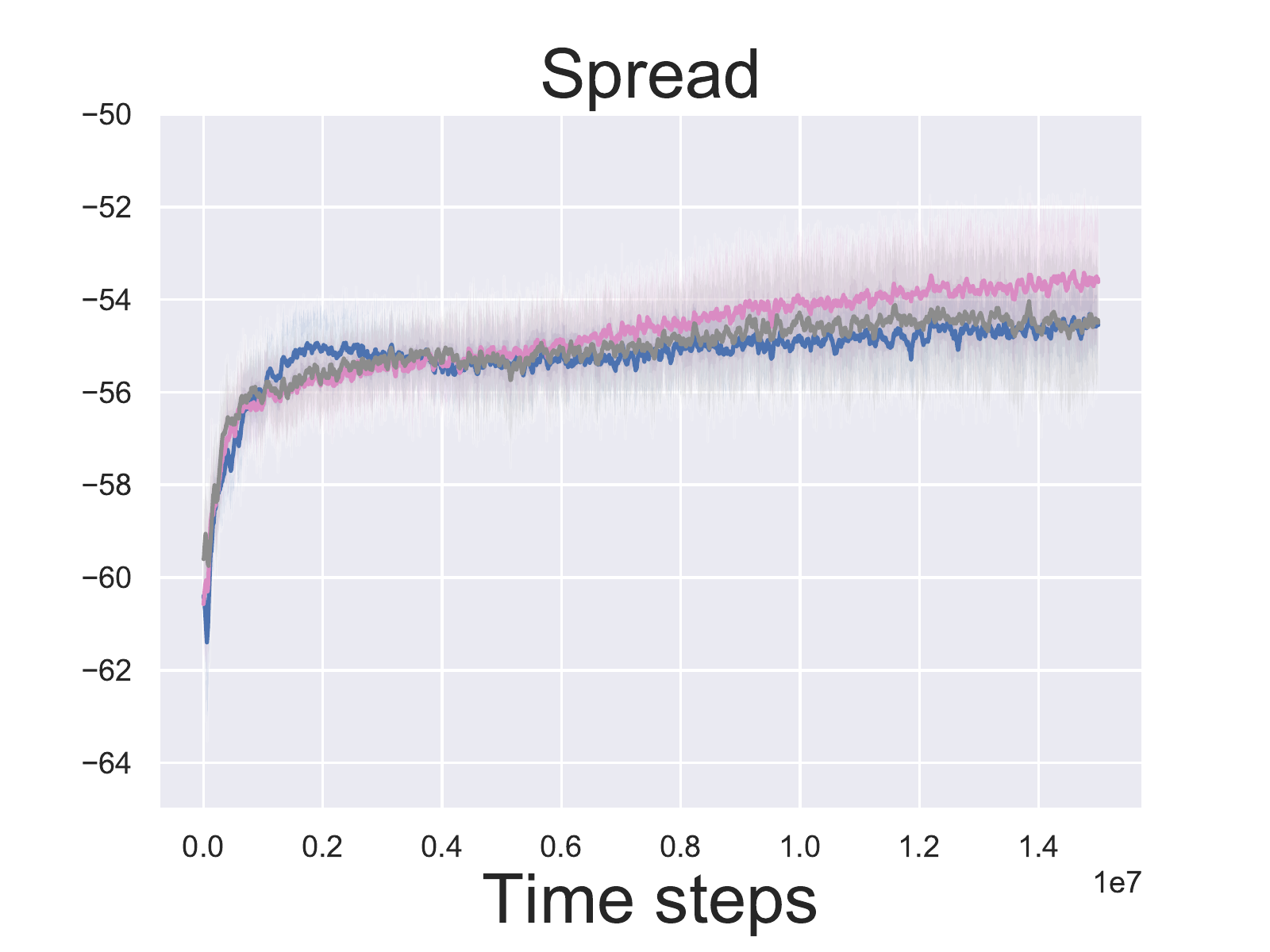}
  \label{fig:double_speaker_listener_srl_disc_abl}
\end{subfigure}\hfil 
\fcolorbox{background_sns}{background_sns}
{ \smac \hspace{0.1cm}  SMA2C Full \hspace{1cm} \smacsa \hspace{0.1cm} SMA2C Observation-Action \hspace{1cm} \smacs \hspace{0.1cm} SMA2C Observation
}
\caption{Ablation on the episodic returns for different inputs in the VAE of SMA2C for weak (top row) and strong (bottom row) generalization in all four environments.}
\label{fig:rl_perf_abl}
\end{figure*}

\begin{figure*}
    \centering 
\begin{subfigure}{0.25\textwidth}
  \includegraphics[width=\linewidth]{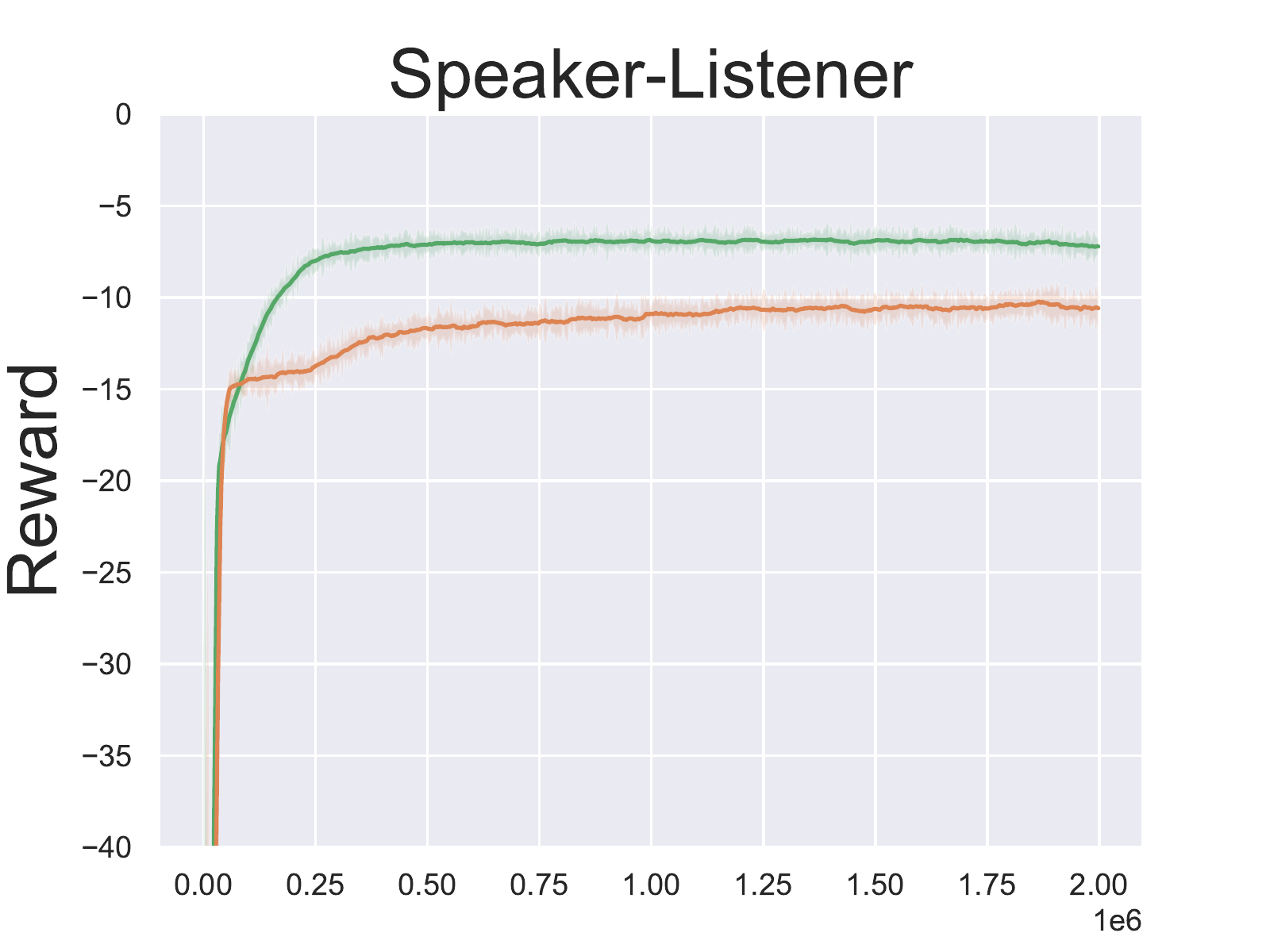}
  \label{fig:speaker_listener_wrl_disc_abl}
\end{subfigure}\hfil 
\begin{subfigure}{0.25\textwidth}
  \includegraphics[width=\linewidth]{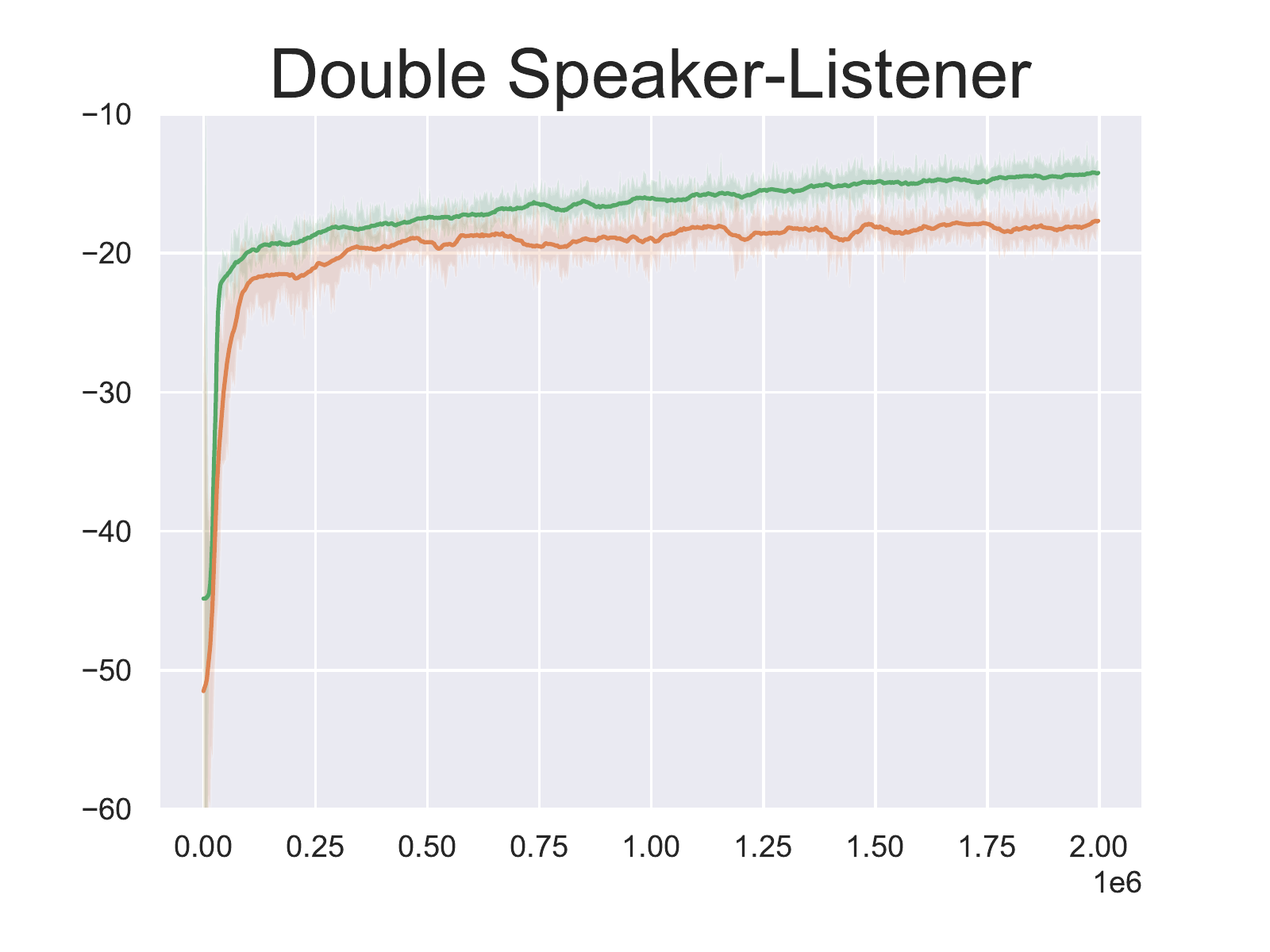}
  \label{fig:speaker_listener_srl_disc_abl}
\end{subfigure}\hfil 
\begin{subfigure}{0.25\textwidth}
 \includegraphics[width=\linewidth]{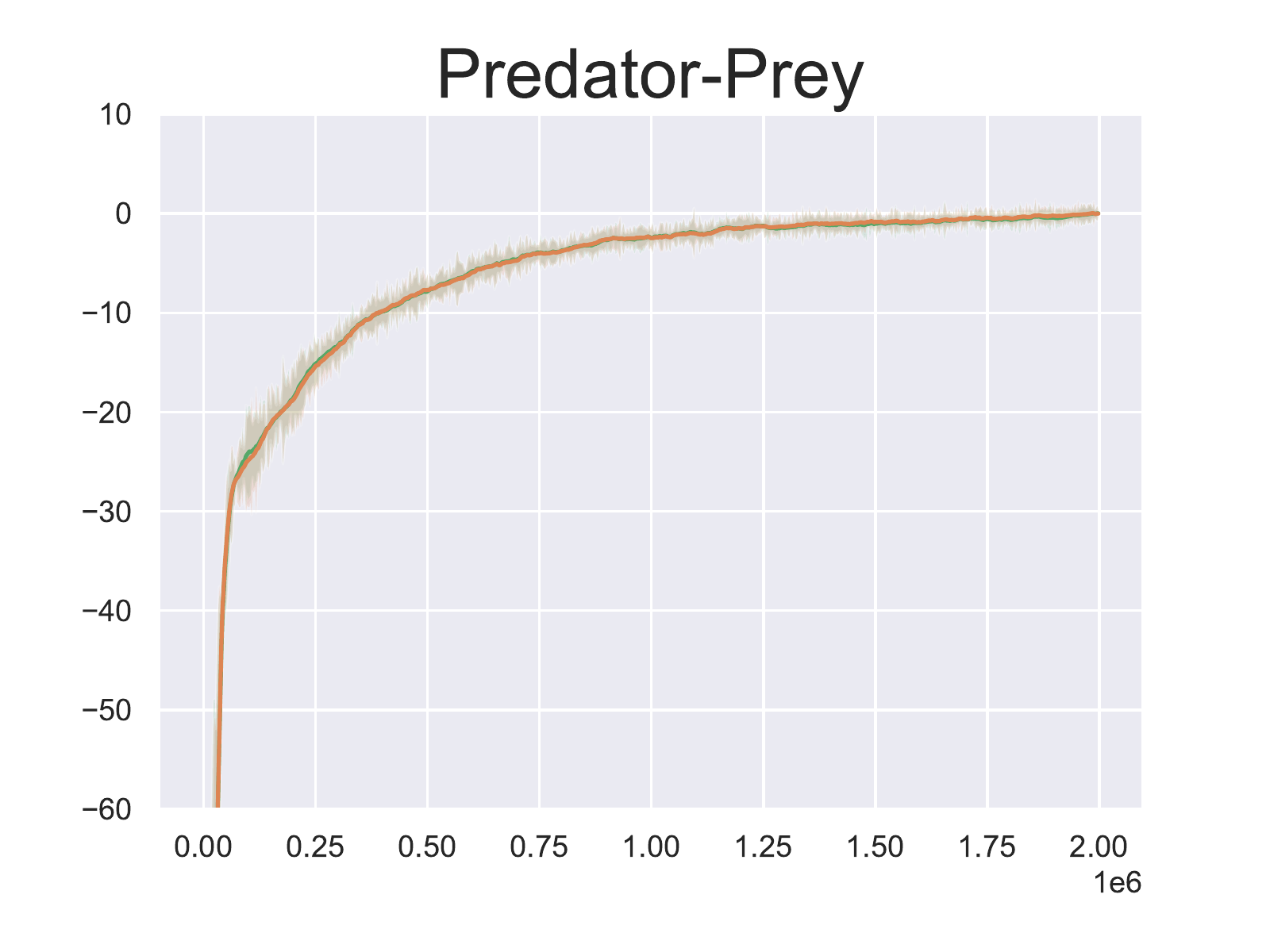}
  \label{fig:double_speaker_listener_wrl_disc_abl}
\end{subfigure}\hfil 
\begin{subfigure}{0.25\textwidth}
  \includegraphics[width=\linewidth]{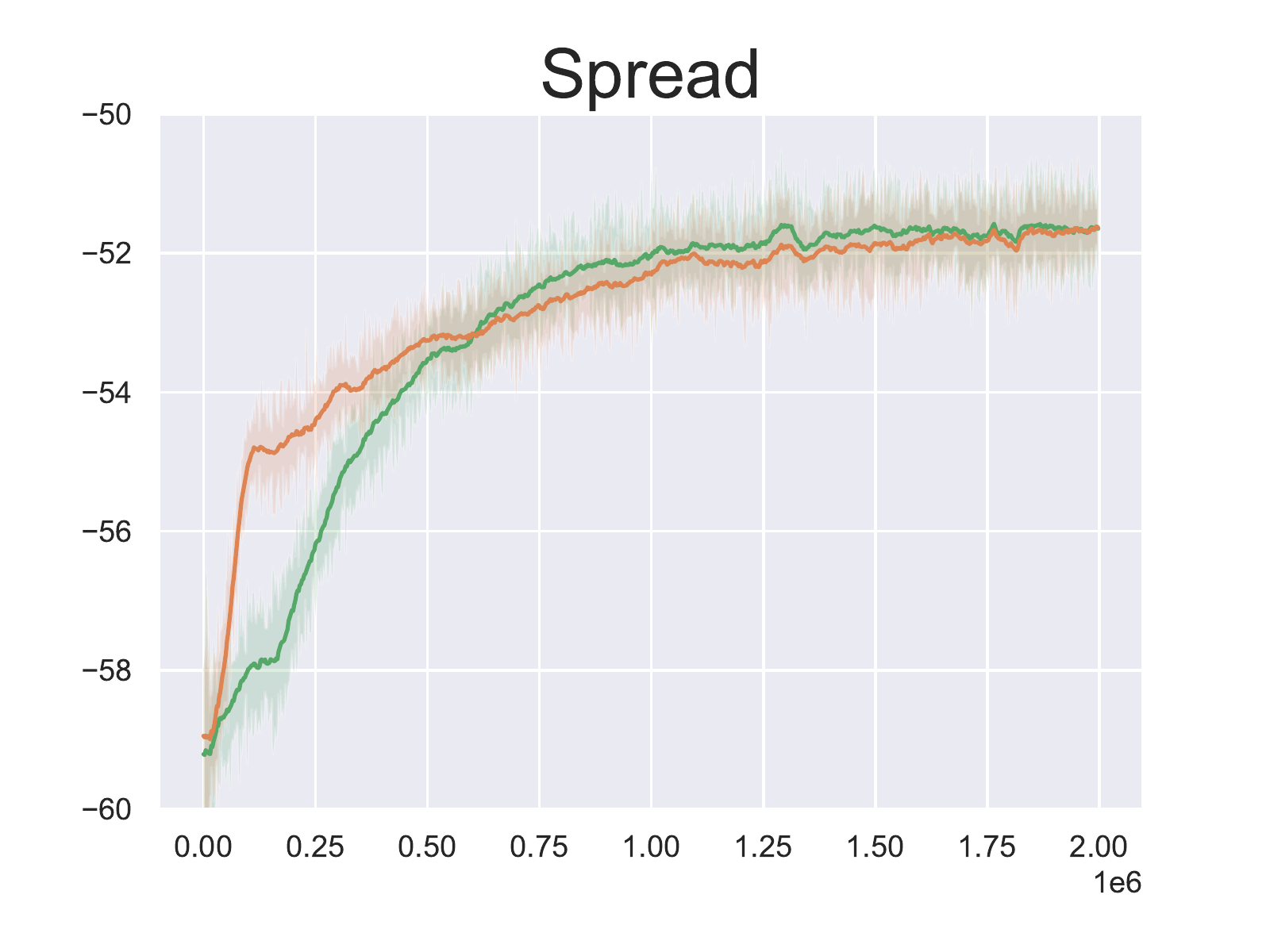}
  \label{fig:double_speaker_listener_srl_disc_abl}
\end{subfigure}\hfil 

    \centering 
\begin{subfigure}{0.25\textwidth}
  \includegraphics[width=\linewidth]{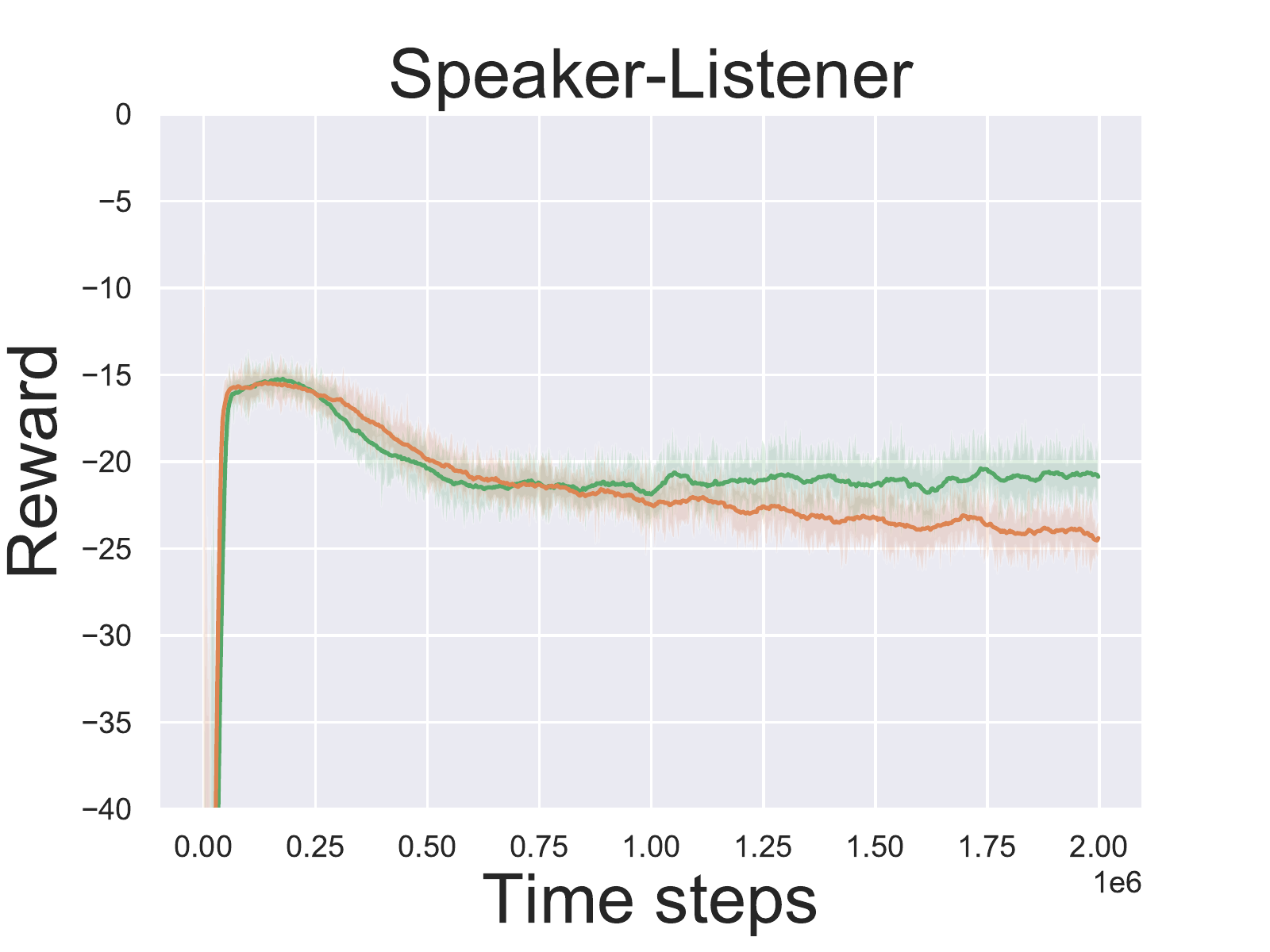}
  \label{fig:speaker_listener_wrl_disc_abl}
\end{subfigure}\hfil 
\begin{subfigure}{0.25\textwidth}
  \includegraphics[width=\linewidth]{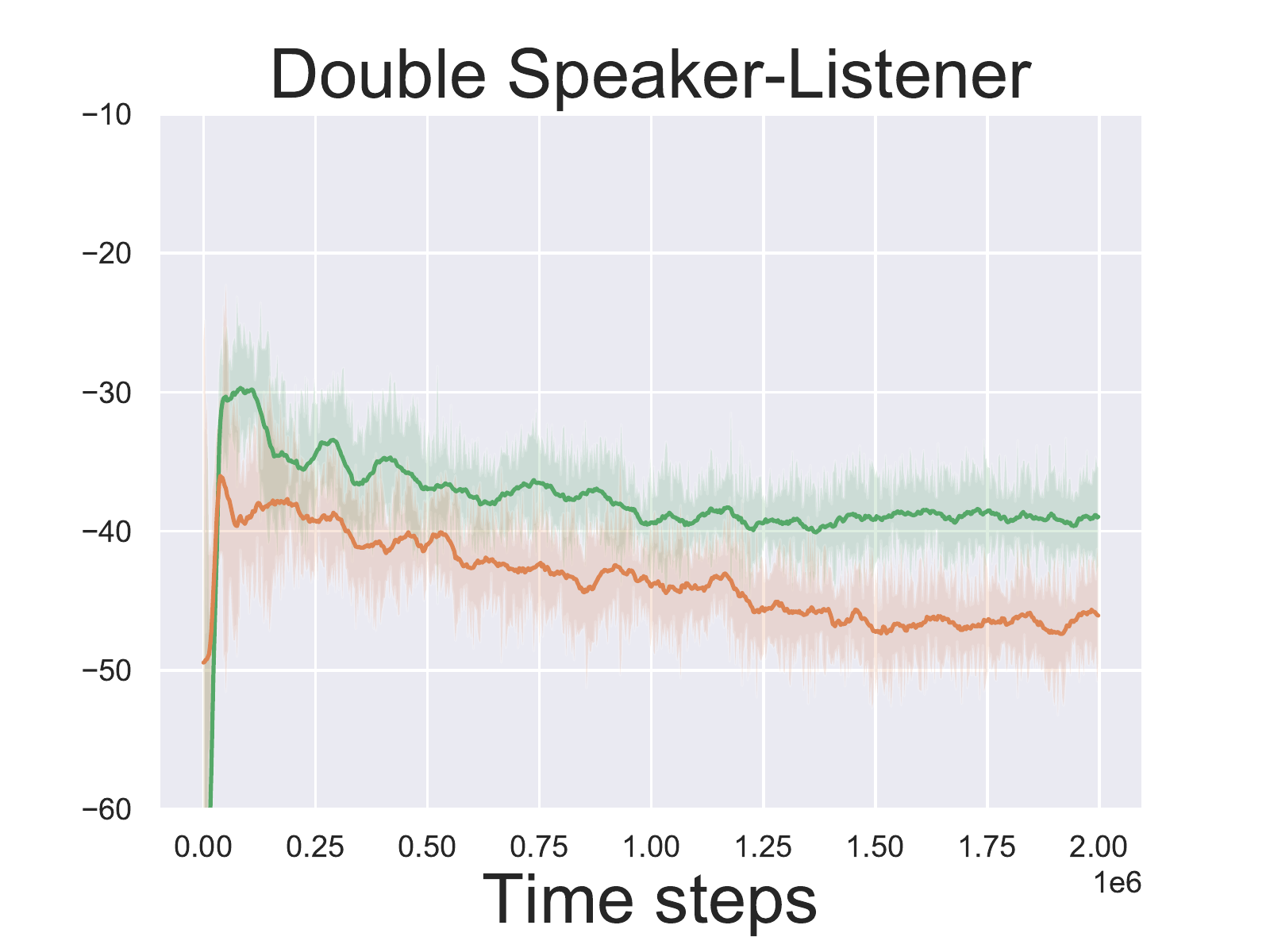}
  \label{fig:speaker_listener_srl_disc_abl}
\end{subfigure}\hfil 
\begin{subfigure}{0.25\textwidth}
 \includegraphics[width=\linewidth]{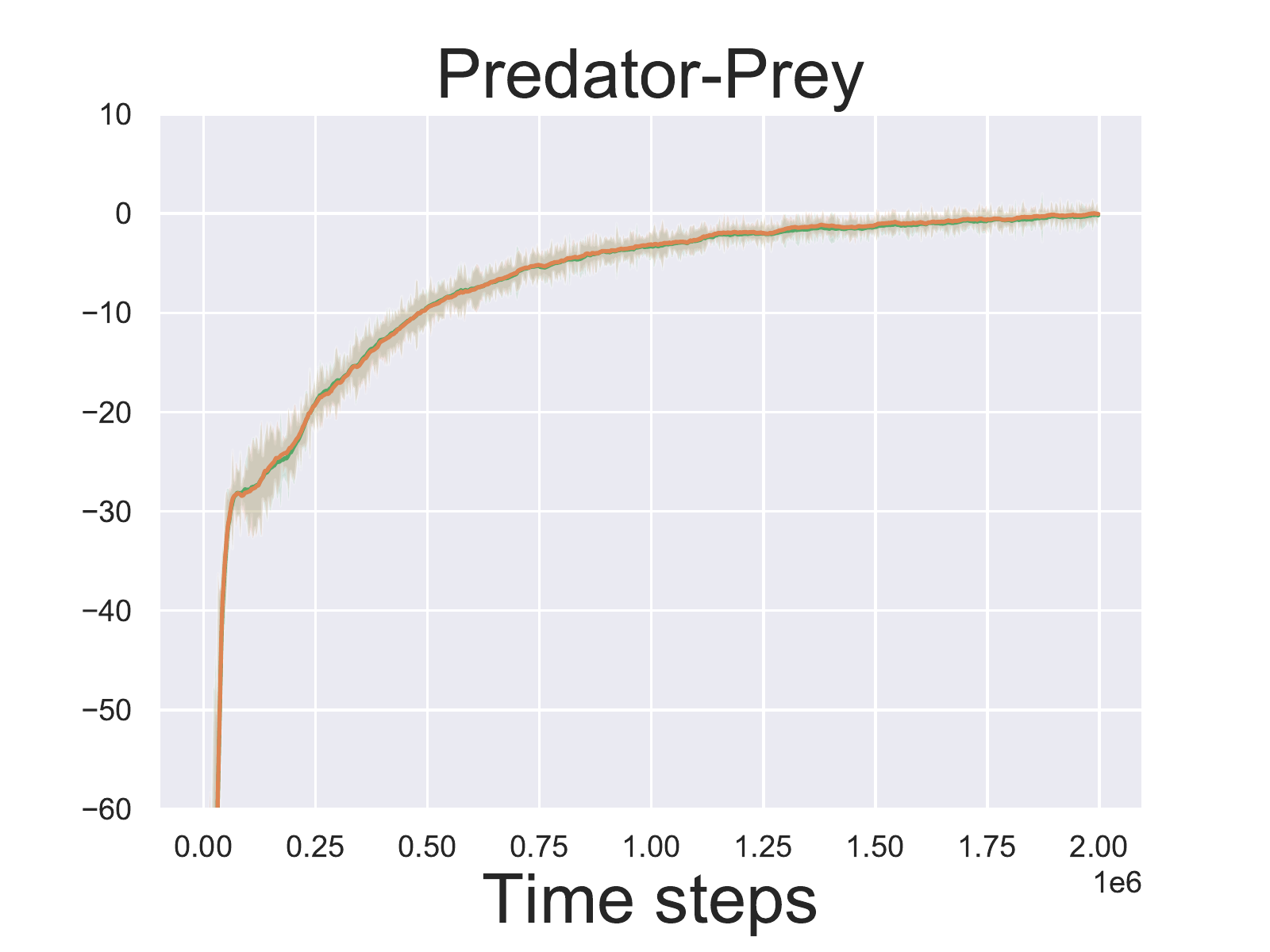}
  \label{fig:double_speaker_listener_wrl_disc_abl}
\end{subfigure}\hfil 
\begin{subfigure}{0.25\textwidth}
  \includegraphics[width=\linewidth]{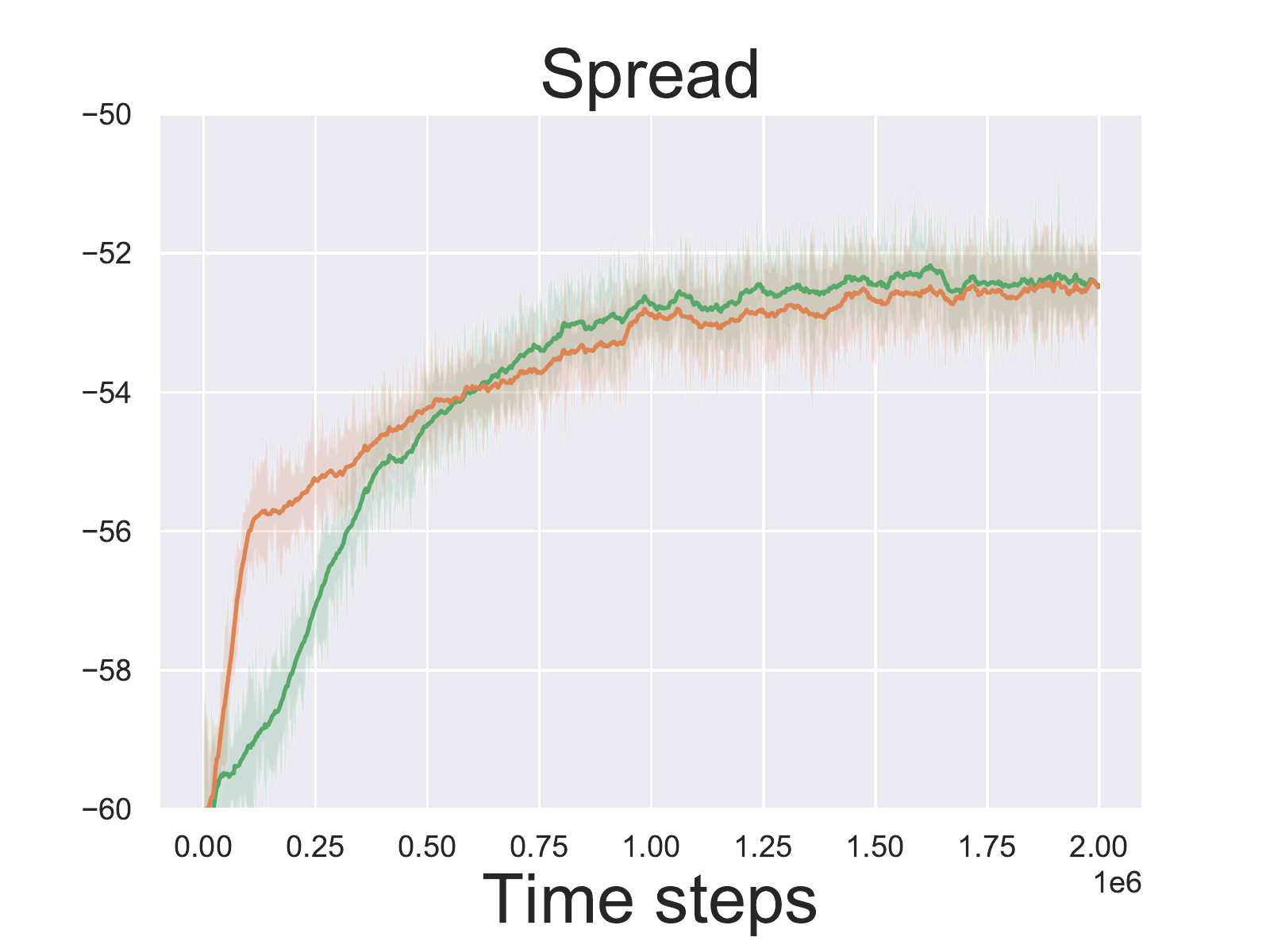}
  \label{fig:double_speaker_listener_srl_disc_abl}
\end{subfigure}\hfil 
\fcolorbox{background_sns}{background_sns}
{ \omddpg \hspace{0.1cm}  OMDDPG \hspace{1cm} \omddpgnodisc \hspace{0.1cm} OMDDPG No Discrimination
}
\caption{Ablation study on the importance of the discrimination objective. Episodic returns in all four environments for weak (top row) and strong (bottom row) generalization.}
\label{fig:rl_discr_abl}
\end{figure*}

We evaluate the proposed opponent modeling methods in RL settings. In Figure \ref{fig:rl_perf}, the episodic returns for the three methods in all four environments are presented.

Every line corresponds to the average return over five runs with different initial seeds, and the shadowed part represents the $95\%$ confidence interval. We evaluate the models every 1000 episodes for 100 episodes. During the evaluation, we sample an embedding from the variational distribution at each time step, and the agent follows the greedy policy. The hyperparameters for all the experiments in Figure \ref{fig:rl_perf} were optimized on weak generalization scenarios, against opponents from the set $\sT$. 

OMDDPG is an upper baseline for SMA2C achieving higher returns in all environments during weak generalization. However, OMDDPG, as well as the method by \citet{grover2018learning}, tend to overfit and perform poorly during strong generalization in the speaker-listener and double speaker-listener environment. SMA2C achieves higher returns than the method proposed by \citet{grover2018learning} in more than half of the scenarios. 

\subsection{Evaluation of the Representations}
\label{sec:rep_eval}

To evaluate the representations created from our models we will estimate the Mutual Information (MI) between the variational distribution ($q(\vz | \vtau)$ or $q(\vz | \vtau_{-1})$) and the prior on the opponents' identities, which is uniform. This is a common way to estimate the quality of the representation \citep{chen2018isolating, hjelm2018learning}. To estimate the MI, we use the Mutual Information Neural Estimation (MINE) \citep{belghazi2018mine}. Note that the upper bound of the MI, the entropy of the uniform distribution, in our experiments is $1.61$. We gather $200$ trajectories against each opponent in $\sT$, where $80\%$ of them are used for training and the rest for testing.

From Table \ref{tab:mine_valus}, we observe that the method of \citet{grover2018learning} achieves significantly higher values of MI. We believe that the main reason behind this is the discrimination objective that implicitly increases MI. This is apparent in the MI values of OMDDPG as well. SMA2C manages to create opponent representations, based only on the local information of our agent, that have statistical dependency with the opponent identities. Additionally, based on Figure \ref{fig:rl_perf}, we observe that the value of MI  does not directly correlate to the episodic returns in RL tasks.

\subsection{Ablation Study on SMA2C Inputs}
\label{sec:abl_sma2c}
We perform an ablation study to assess the performance requirements of SMA2C. Our proposed method utilizes the observation, action, reward, and termination sequence of our agent to generate the opponent's model. We use different combinations of inputs in the encoder and compare the episodic returns. In Figure \ref{fig:rl_perf_abl}, the average episode return is presented for three different cases; SMA2C, SMA2C with only observations and actions and SMA2C with only observations. From Figure \ref{fig:rl_perf_abl}, we observe that during weak generalization the performance of SMA2C without using the reward and the termination variable is comparable to SMA2C in all environments. However, during strong generalization the performance SMA2C without using the reward and the termination variable drops in the speaker-listener and double speaker-listener environment. SMA2C using only the observations for modeling has similar performance to SMA2C that uses both the observations and actions of our agent in all environments except the predator-prey.

\subsection{Ablation Study on Discrimination Objective}
\label{sec:abl_disc}
Another element of this work is the utilization of the discrimination objective proposed by \citet{grover2018learning} in the VAE loss. To better understand how the opponent separation in the embedding space is related to RL performance. Figure \ref{fig:rl_discr_abl} shows the episodic return during the training for the OMDDPG with and without the discrimination objective. Using the discrimination objective has a significant impact on the episodic returns in the speaker-listener and the double speaker-listener environment. From Figure \ref{fig:rl_discr_abl} we observe that the discrimination objective increases the performance of OMDDPG in the speaker-listener and double speaker-listener environment for weak and strong generalization.

\section{Conclusion}

To conclude this work, we proposed two methods for opponent modeling in multi-agent systems using variational autoencoders. First, we proposed OMDDPG, a VAE-based method that uses the common assumption that access to opponents' information is available during execution. The goal of this work is to motivate opponent modeling without access to opponent's information. The core contribution of this work is SMA2C, which learns representations without requiring access to opponent's information during execution. To the best of our knowledge this is the first study showing that effective opponent modeling can be achieved without requiring access to opponent observations.
In the future, we would like to research how these models can be used for non-stationary opponents. In particular, we plan on investigating two scenarios; the first is multi-agent deep RL, where different agents are learning concurrently leading to non-stationarity in the environment \citep{papoudakis2019dealing}, which prevents the agents from learning optimal policies. Secondly, we would like to explore whether the proposed models can deal with opponents that try to deceive it and exploit the controlled agent \citep{ganzfried2011game, ganzfried2015safe}.

\bibliography{aaai_2020}
\end{document}